\PassOptionsToPackage{table}{xcolor}
\documentclass{article} 

\usepackage{iclr2025_conference,times}

\usepackage{wrapfig}

\usepackage{booktabs}      
\usepackage{multirow}      
\usepackage{amssymb}       
\usepackage{graphicx}     

\usepackage{colortbl}
\usepackage{xcolor}

\usepackage{hyperref}

\usepackage{url}
\usepackage{multicol}
\usepackage{bm}
\usepackage{aliascnt}
\usepackage{adjustbox}
\usepackage{enumitem}
\usepackage{amsmath}  
\usepackage{longtable} 
\usepackage{subcaption}
\usepackage{calc}
\usepackage[most]{tcolorbox}
\usepackage{listings}
\usepackage{caption}

\definecolor{uclablue}{rgb}{0.15, 0.45, 0.68}
\hypersetup{
    breaklinks,
    citecolor=uclablue,
    colorlinks=true,
}

\newtcolorbox{AIbox}[2][]{aibox,title=#2,#1}
\tcbset{
  aibox/.style={
    top=10pt,
    colframe=black,
    colbacktitle=black,
    coltitle=white,
    center,
  }
}

\lstdefinelanguage{prompt}{
    basicstyle=\scriptsize\ttfamily, 
    mathescape=true,        
    escapebegin=\color{latentcolor},  
    escapeend={},
    escapechar=@,
    stringstyle = \color{myorange},
    showstringspaces = false,
    moredelim = [s][\color{mypink}]{`}{`},
    moredelim = [s][\color{mybrown}]{```json}{```},
    moredelim = [s][\color{latentcolor}]{<StartOfLatent>}{<EndOfLatent>},
    literate = %
        {\ \ a.\ }{{\textcolor{mypurple}{\ \ a.\ }}}5
        {\ \ b.\ }{{\textcolor{mypurple}{\ \ b.\ }}}5
        {\ \ c.\ }{{\textcolor{mypurple}{\ \ c.\ }}}5
        {\ \ d.\ }{{\textcolor{mypurple}{\ \ d.\ }}}5
        {\ \ e.\ }{{\textcolor{mypurple}{\ \ e.\ }}}5
        {\ \ f.\ }{{\textcolor{mypurple}{\ \ f.\ }}}5
        {\ \ g.\ }{{\textcolor{mypurple}{\ \ g.\ }}}5
        {\ \ h.\ }{{\textcolor{mypurple}{\ \ h.\ }}}5
        {\ I.\ }{{\textcolor{mypurple}{\ I.\ }}}4
        {\ II.\ }{{\textcolor{mypurple}{\ II.\ }}}5
        {\ III.\ }{{\textcolor{mypurple}{\ III.\ }}}6
        {\ IV.\ }{{\textcolor{mypurple}{\ IV.\ }}}5
        {\ V.\ }{{\textcolor{mypurple}{\ V.\ }}}4
}

\newtcblisting[auto counter, number within=subsection]{prompt}[4][]{%
  before skip=6pt, after skip=6pt,    
  top=6pt, bottom=6pt, left=7pt, right=7pt, 
  enhanced, breakable,
  colback=white, colframe=gray!65,
  boxrule=0.6pt, arc=2pt,
  drop shadow={black!15},             
  listing only,
  listing options={
    language=prompt,
    upquote=true,
    basicstyle=\scriptsize\ttfamily \setlength{\baselineskip}{1.1\baselineskip},
    columns=fullflexible,
    breaklines=true,
    breakindent=0pt,
    xleftmargin=\ifx\empty#2\empty0pt\else#2\fi,
    xrightmargin=\ifx\empty#3\empty0pt\else#3\fi,
    aboveskip=0pt,                    
    belowskip=0pt,                    
    showstringspaces=false,
  },
  title={\ifstrempty{#4}{Prompt \thetcbcounter}{Prompt \thetcbcounter: #4}},
  colbacktitle=gray!6, coltitle=black,
  attach boxed title to top left={yshift=-1mm, xshift=6pt},
  boxed title style={colback=gray!6, boxrule=0pt, sharp corners},
  #1
}

\newtcblisting[auto counter, number within=subsection]{showcase}[4][]{%
  before=\par\vspace{\baselineskip},
  after=\par,
  width=\linewidth,
  enhanced,
  arc=0em,
  boxrule=1pt,
  listing only,
  listing options={
    language=prompt,
    upquote=true,
    basicstyle=\scriptsize\ttfamily \setlength{\baselineskip}{1.1\baselineskip},
    breaklines=true,
    breakindent=0pt,
    xleftmargin=\ifx\empty#2\empty-12pt\else#2\fi,
    xrightmargin=\ifx\empty#3\empty-5pt\else#3\fi,
    aboveskip=-4pt,
    belowskip=-4pt,
    columns=fullflexible,
    },
  colback=white,
  colframe=gray,
  colbacktitle=gray!5,
  coltitle=black,
  attach boxed title to top center={yshift=-3mm},
  box align=center,
  parbox=false,
  title={\ifstrempty{#4}{Example \thetcbcounter}{Example \thetcbcounter: #4}},
  #1
}

\definecolor{linkColor}{rgb}{0.2,0.4,0.6}
\definecolor{myblue}{HTML}{0379AC}
\definecolor{myred}{HTML}{A50E50}
\definecolor{myorange}{RGB}{238, 133, 74}
\definecolor{latentcolor}{named}{cyan}
\definecolor{normalcolor}{RGB}{0, 0, 0}
\usepackage{marvosym}
\definecolor{lightblue1}{rgb}{0.97, 0.985, 1} 
\definecolor{lightblue2}{rgb}{0.92, 0.965, 1} 
\definecolor{lightblue3}{rgb}{0.84, 0.93, 1}
\definecolor{lightblue4}{rgb}{0.74, 0.87, 1}
\definecolor{lightblue5}{rgb}{0.64, 0.81, 1}
\definecolor{lightblue6}{rgb}{0.54, 0.75, 1}

\definecolor{lightgreen1}{rgb}{0.97, 1.00, 0.97}
\definecolor{lightgreen2}{rgb}{0.92, 0.98, 0.92}
\definecolor{lightgreen3}{rgb}{0.84, 0.95, 0.84}
\definecolor{lightgreen4}{rgb}{0.74, 0.91, 0.74}
\definecolor{lightgreen5}{rgb}{0.64, 0.86, 0.64}
\definecolor{lightgreen6}{rgb}{0.54, 0.81, 0.54}

\definecolor{lightorange1}{rgb}{1.00, 0.98, 0.95}
\definecolor{lightorange2}{rgb}{1.00, 0.95, 0.85}
\definecolor{lightorange3}{rgb}{1.00, 0.90, 0.70}
\definecolor{lightorange4}{rgb}{1.00, 0.85, 0.55}
\definecolor{lightorange5}{rgb}{1.00, 0.80, 0.40}
\definecolor{lightorange6}{rgb}{1.00, 0.75, 0.30}

\definecolor{lightpurple1}{rgb}{0.985, 0.97, 1.00}
\definecolor{lightpurple2}{rgb}{0.96, 0.92, 1.00}
\definecolor{lightpurple3}{rgb}{0.93, 0.84, 1.00}
\definecolor{lightpurple4}{rgb}{0.87, 0.74, 1.00}
\definecolor{lightpurple5}{rgb}{0.81, 0.64, 1.00}
\definecolor{lightpurple6}{rgb}{0.75, 0.54, 1.00}

\definecolor{lightred1}{rgb}{1.00, 0.97, 0.97}
\definecolor{lightred2}{rgb}{1.00, 0.92, 0.92}
\definecolor{lightred3}{rgb}{1.00, 0.84, 0.84}
\definecolor{lightred4}{rgb}{1.00, 0.74, 0.74}
\definecolor{lightred5}{rgb}{1.00, 0.64, 0.64}
\definecolor{lightred6}{rgb}{1.00, 0.54, 0.54}

\definecolor{lightcyan1}{rgb}{0.97, 1.00, 1.00}
\definecolor{lightcyan2}{rgb}{0.92, 0.98, 0.98}
\definecolor{lightcyan3}{rgb}{0.84, 0.95, 0.96}
\definecolor{lightcyan4}{rgb}{0.74, 0.91, 0.94}
\definecolor{lightcyan5}{rgb}{0.64, 0.87, 0.92}
\definecolor{lightcyan6}{rgb}{0.54, 0.83, 0.90}

\definecolor{Gray}{gray}{0.85}
\definecolor{LightCyan}{rgb}{0.88,1,1}
\definecolor{greyC}{RGB}{180,180,180}
\definecolor{greyL}{RGB}{235,235,235}
\definecolor{citeColor}{RGB}{0,20,115}
\hypersetup{colorlinks,linkcolor={red},citecolor={citeColor},urlcolor={citeColor}}
\definecolor{shadecolor}{rgb}{0.92,0.92,0.92}

\usepackage{cleveref}
\crefname{template}{Template}{Template}

\usepackage{pifont}
\definecolor{rliableblue}{RGB}{0, 102, 204} 

\tcbset{
    myblueboxstyle/.style={
        colback=lightblue3, 
        colframe=lightblue5, 
        coltitle=black, 
        fonttitle=\bfseries, 
        boxrule=1pt, 
        width=\linewidth, 
        left=2pt, 
        right=2pt, 
        top=2pt, 
        bottom=2pt, 
        sharp corners, 
        boxsep=2pt
    }
}

\lstdefinestyle{iclrstyle}{
    language=Python,
    basicstyle=\ttfamily\small,  
    columns=fullflexible,        
    keepspaces=true,             
    showspaces=false,            
    showstringspaces=false,      
    commentstyle=\color{gray}\itshape, 
    keywordstyle=\color{codekw}\bfseries, 
    stringstyle=\color{myorange}, 
    escapechar=|,                
    frame=none,                  
    xleftmargin=1.5em,           
    aboveskip=0.5em,             
    belowskip=0.5em,             
    breaklines=true,             
    breakindent=0pt,
}

\usepackage{float}
\newcommand{\xmark}{\textcolor{red!70!black}{\ding{55}}}   
\newcommand{\cmark}{\textcolor{green!60!black}{\ding{51}}} 

\usepackage{titlesec}
\definecolor{objblue}{RGB}{3,139,221}  
\definecolor{attrred}{RGB}{255,67,67}    
\definecolor{easygreen}{RGB}{0,156,75}  
\definecolor{middleyellow}{RGB}{242,89,34}  
\definecolor{hardred}{RGB}{216,56,58}
\usepackage{array}
\definecolor{BoxBackground}{RGB}{240, 240, 240} 
\definecolor{BoxFrame}{RGB}{0, 0, 0} 
\definecolor{TitleBackground}{RGB}{0, 0, 0} 
\definecolor{TitleText}{RGB}{255, 255, 255} 
\tcbset{
  academicbox/.style={
    boxsep=5pt,
    left=2pt,
    right=2pt,
    bottom=0.5pt,
    boxrule=0.5pt,
    colback=BoxBackground,
    colframe=BoxFrame,
    colbacktitle=TitleBackground,
    coltitle=TitleText,
    enhanced,
    attach boxed title to top left={yshift=-0.1in,xshift=0.1in},
    boxed title style={boxrule=0pt,colframe=white},
    title={#1},
  }
}
\newtcolorbox{AcademicBox}[1][]{academicbox=#1}

\usepackage{algorithm}
\usepackage{algpseudocode}
\usepackage{etoolbox}
\makeatletter
\AfterEndEnvironment{algorithm}{\let\@algcomment\relax}
\AtEndEnvironment{algorithm}{\kern2pt\hrule\relax\vskip3pt\@algcomment}
\let\@algcomment\relax
\newcommand\algcomment[1]{\def\@algcomment{\footnotesize#1}}
\renewcommand\fs@ruled{\def\@fs@cfont{\bfseries}\let\@fs@capt\floatc@ruled
  \def\@fs@pre{\hrule height.8pt depth0pt \kern2pt}%
  \def\@fs@post{}%
  \def\@fs@mid{\kern2pt\hrule\kern2pt}%
  \let\@fs@iftopcapt\iftrue}
\makeatother

\NewDocumentCommand{\xx}
{ mO{} }{\textcolor{blue}{\textsuperscript{\textit{todo}}\textsf{\textbf{\small[#1]}}}}
\usepackage{xspace}

\definecolor{codeblue}{rgb}{0.25,0.5,0.5}
\definecolor{codekw}{rgb}{0.85, 0.18, 0.50}
\definecolor{diffgreen}{rgb}{0.0, 0.6, 0.0} 
\definecolor{diffred}{rgb}{0.8, 0.0, 0.0}   

\title{OmniWeaving: Towards Unified Video Generation with Free-form Composition and Reasoning} 

\author{
Kaihang Pan$^{1,2, *, \ddagger}$, 
Qi Tian$^{2, *}$,  
Jianwei Zhang$^{2}$, 
Weijie Kong$^{2}$, 
Jiangfeng Xiong$^{2}$, \\
Yanxin Long$^{2}$, 
Shixue Zhang$^{2}$, 
Haiyi Qiu$^{1}$, 
Tan Wang$^{3}$, 
Zheqi Lv$^{1}$,   
Shuyuan Tu$^{2}$, \\
Yue Wu$^{2, \S}$, 
Liefeng Bo$^{2}$, 
Siliang Tang$^{1, \S}$, 
Zhao Zhong$^{2,\dagger}$\\
\textbf{$^1$Zhejiang University}  \textbf{$^2$Tencent Hunyuan}  \textbf{$^3$Nanyang Technological University}\\
$^*$ Equal Contribution, $^\S$ Corresponding Authors, $^\dagger$  Project Leader \\
{\ttfamily\normalsize 
\raisebox{-1.5pt}{\includegraphics[height=1.05em]{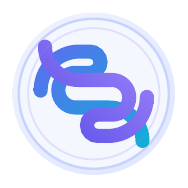}} Project Page: \url{https://omniweaving.github.io/} \\
\raisebox{-1.5pt}{\includegraphics[height=1.0em]{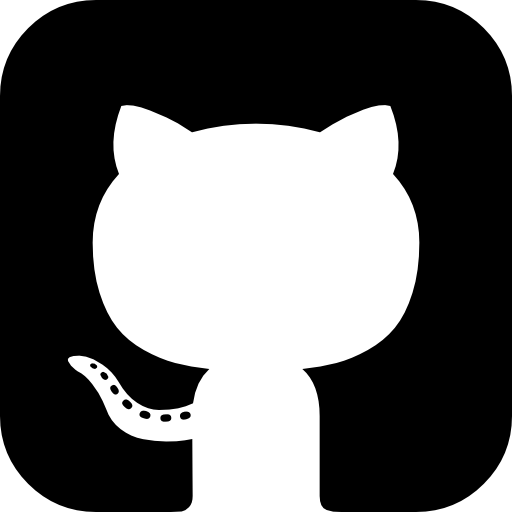}} GitHub: \url{https://github.com/Tencent-Hunyuan/OmniWeaving} \\
\raisebox{-1.5pt}{\includegraphics[height=1.0em]{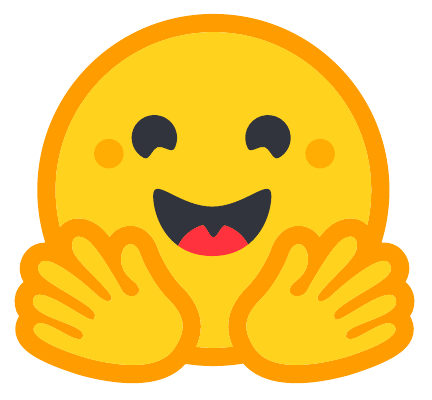}} Model: \url{https://huggingface.co/tencent/HY-OmniWeaving} 
}
}

\begin{document}
\maketitle
\renewcommand*{\thefootnote}{\fnsymbol{footnote}}
\footnotetext{$^\ddagger$ Work done when interning at Tencent Hunyuan.}

\begin{abstract}
  While proprietary systems such as Seedance-2.0 have achieved remarkable success in omni-capable video generation, the academic research community lags far behind: most of its models remain heavily fragmented, and the few existing efforts toward unified video generation still struggle to seamlessly integrate diverse tasks within a single framework. To bridge this gap, we propose OmniWeaving, an omni-level video generation model featuring powerful multimodal composition and reasoning-informed capabilities. By leveraging a massive-scale pretraining dataset that encompasses diverse compositional and reasoning-augmented scenarios, OmniWeaving learns to temporally bind interleaved text, multi-image, and video inputs while acting as an intelligent agent to infer complex user intentions for sophisticated video creation. Furthermore, we introduce IntelligentVBench, the first comprehensive benchmark designed to rigorously assess next-level intelligent unified video generation. Extensive experiments demonstrate that OmniWeaving achieves SoTA performance among open-source academic unified models. The code and model are publicly available. Project Page: \url{https://omniweaving.github.io}.
\end{abstract}
\section{Introduction}

\begin{figure}[htbp]
    \centering
    \includegraphics[width=\textwidth]{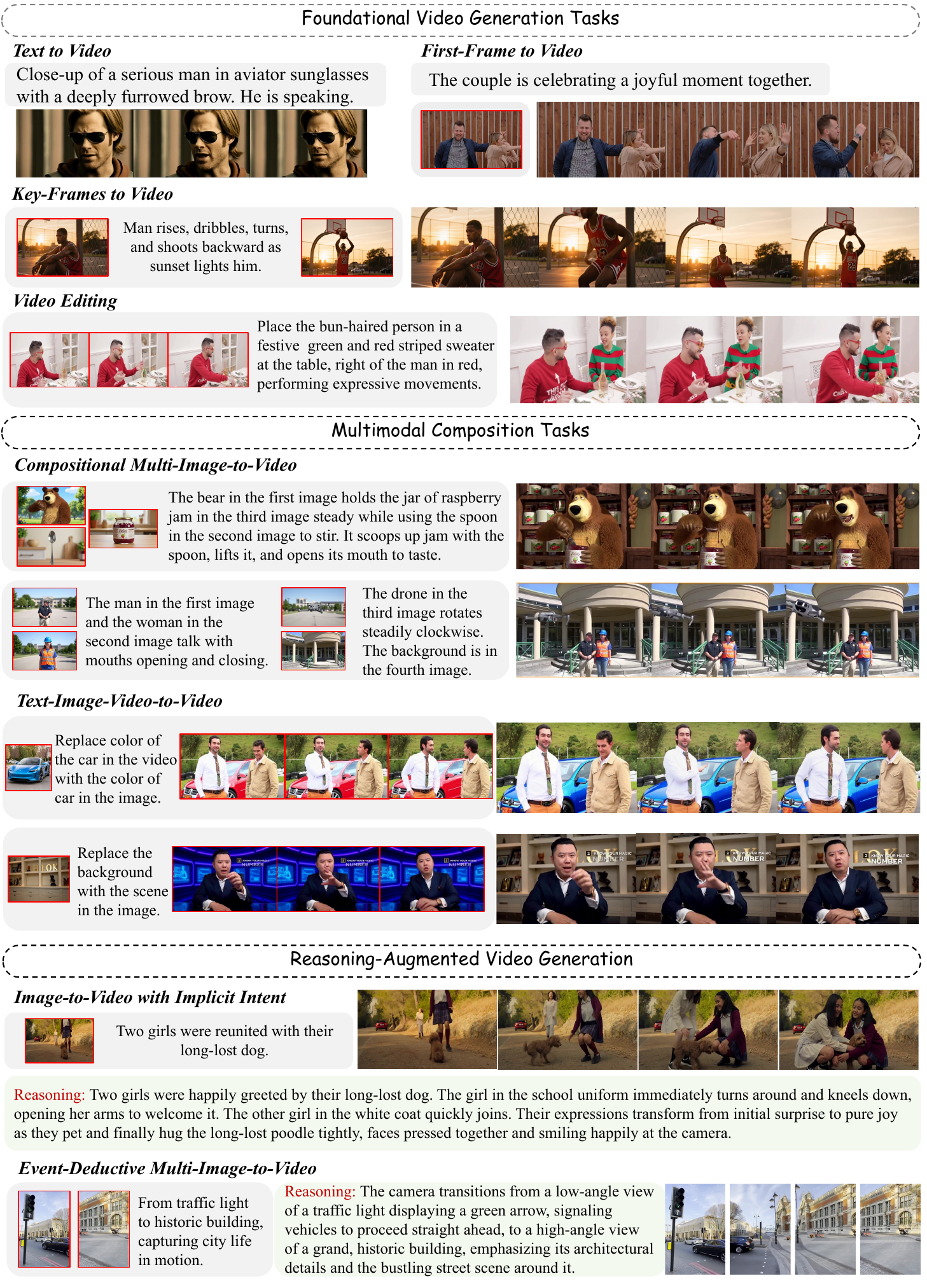} 
    \caption{Showcase of OmniWeaving across diverse video generation scenarios, such as foundational  tasks, multimodal composition tasks, and reasoning-augmented scenarios.}
    \label{fig:intro}
\end{figure}

The pursuit of artificial general intelligence has driven the evolution of visual generation models from task-specific experts to unified generalists~\citep{openai2024gpt4ocard,xiao2025omnigen,pan2025generative,pan2024auto,xia2025dreamomni}. 
In the image domain, this paradigm shift was significantly catalyzed by GPT-4o~\citep{openai2024gpt4ocard} and NanoBanana~\citep{gemini2_5_flash_imag}, proprietary models that seamlessly integrated image understanding and generation within a single framework. 
Their unprecedented success in executing omni-level generation has sparked a vigorous response from the open-source community.
Consequently, academic models like BAGEL~\citep{deng2025emerging} and OmniGen2~\citep{wu2025omnigen2} have rapidly emerged, natively coupling visual comprehension with generative modules to enable unified image synthesis with free-form multimodal inputs.

As the field naturally progresses toward the temporal domain, video generation also reaches a pivotal juncture requiring a more unified framework. Recently, industrial systems such as Seedance-2.0~\citep{seedance2.0} have redefined the landscape, establishing that next-generation models must be genuinely ``omni-capable'' by synergizing two foundational pillars: \textbf{(1) Multimodal composition}, which enables the seamless spatio-temporal binding of free-form, interleaved text, image, and video inputs; and \textbf{(2) Abstract reasoning}, which empowers models to act as active agents capable of inferring complex user intentions and mastering the underlying semantic logic of dynamic scenes.

However, unlike the flourishing academic ecosystem in image generation, its counterpart in unified video generation significantly lags behind industrial systems, revealing a substantial capability gap. 
First, the current landscape of video generation remains dominated by fragmented approaches~\citep{wu2025hunyuanvideo, wan2025wan, kong2024hunyuanvideo} narrowly tailored for text-to-video, image-to-video synthesis, or video-to-video synthesis~\citep{he2025openve}, relying on task-specific modules that impede scaling and integration.
Furthermore, while recent academic models such as VACE~\citep{jiang2025vace}, UniVideo~\citep{wei2025univideo}, and VINO~\citep{chen2026vino} attempt to unify video generation tasks, they either focus primarily on basic task combinations or fail to leverage deep visual understanding to drive unified generation. Consequently, they still struggle to effectively address multimodal composition and reasoning-informed video synthesis.

We argue that bridging this substantial capability gap of unified video models relies on three key drivers. \textbf{First}, the model architecture must integrate both visual comprehension and generation into a single framework to explicitly activate abstract reasoning, evolving models from passive renderers into ``thinking-guided'' generators. 
\textbf{Second}, a transition toward free-form, multi-task pretraining is essential to move beyond rigid prompt-video pairs and capture the intricate semantic relationships across diverse modalities. 
\textbf{Finally}, since existing benchmarks are largely limited to simplistic  tasks with monolithic input formats, the community requires a more complex and comprehensive evaluation suite to foster the development of truly ``omni-capable'' video systems.

To address these critical challenges, we propose OmniWeaving, an omni-level video generation framework capable of both multimodal composition and abstract reasoning. 
Based on a unified architecture integrating visual comprehension and generation, we introduce a massive-scale training dataset that spans a broad spectrum of scenarios and diverse input formats, including both multimodal composition and reasoning-augmented tasks. 
Through a meticulous three-stage training strategy, as shown in Figure~\ref{fig:intro}, OmniWeaving can adeptly handle diverse video generation scenarios, effectively ``weaving'' free-form text, image, and video inputs into a coherent spatio-temporal narrative.

To rigorously evaluate unified video generation from heterogeneous, free-form inputs, we introduce IntelligentVBench, a novel benchmark employing a ``VLM-as-a-Judge'' paradigm to assess abstract reasoning and compositional capabilities across four distinct tasks. Extensive experiments demonstrate that our proposed OmniWeaving framework achieves state-of-the-art performance among existing academic unified models. In summary, our main contributions are threefold:
\begin{itemize}
    \item We propose OmniWeaving, a unified framework that seamlessly integrates visual understanding to achieve omni-level video generation from free-form inputs with strong composition and reasoning capabilities.
    \item We introduce a massive-scale dataset, spanning a broad spectrum of generative scenarios including both composition- and reasoning-related training tasks.
    \item We present IntelligentVBench, the first benchmark dedicated to measuring multimodal composition and abstract reasoning in unified video generation.
\end{itemize}

\section{Related Work}

\label{sec:2}

\textbf{Unified Video Generation.}
Video generation~\citep{kong2024hunyuanvideo, wu2025hunyuanvideo, tu2025stableanimator, tu2024motioneditor, tu2025stableavatar, huang2026va, wan2025wan, ma2025step} has witnessed rapid progress in recent years.
Industrial systems such as Seedance-2.0~\citep{seedance2.0}, Kling-O1~\citep{team2025kling}, MiniMax-H3~\citep{minimax}, and Veo3~\citep{veo3} already span a broad spectrum of video generation tasks within a single model, yet none of them fully discloses its training data or recipe, leaving the research community without a reproducible path toward such capability.
Meanwhile, most open-source video generation models remain narrowly tailored for specific tasks, such as text-to-video, image-to-video~\citep{wan2025wan, wu2025hunyuanvideo}, or video-to-video synthesis~\citep{bai2025scaling, he2025openve}, and typically rely on isolated models and disjointed pipelines.
Furthermore, genuine unified video generation fundamentally relies on robust multimodal composition and abstract reasoning, and recent academic efforts attempting such unification still exhibit notable shortcomings. 
For instance, OmniVideo~\citep{tan2025omni} and OmniVideo2~\citep{yang2026omni} merely incorporate two related video generation capabilities—text-to-video and video editing—into a single framework.
Although models like VACE~\citep{jiang2025vace}, UniVideo~\citep{wei2025univideo}, and VINO~\citep{chen2026vino} expand the variety of supported tasks, they fail to fully leverage deep visual understanding to drive unified generation and lack the cohesive integration of multi-task capabilities within a single architecture.
To address these challenges, OmniWeaving aims to explore better strategies across multiple dimensions, including architecture, data, and training paradigms, to provide a robust reference for next-generation unified video synthesis.

\nocite{leng2026preference, yang2026arcflow}

\textbf{Video Generation Benchmarks.} 
As video generation models rapidly advance, traditional benchmarks struggle to capture their true capabilities due to two primary limitations.
\textbf{(1) Lack of complexity:} Most benchmarks are highly task-specific with rigid input formats. For instance, VBench~\citep{huang2024vbench} and VBench++~\citep{huang2025vbench++} strictly evaluate foundational text- or image-to-video generation, restricted to single-shot scenarios. TGVE+~\citep{singer2024video} and OpenVE-Bench~\citep{he2025openve} focus on Video-to-Video editing tasks. While VACE-Bench~\citep{jiang2025vace} attempts to incorporate various downstream tasks, the input structures still remain inflexible.
\textbf{(2) Lack of comprehensiveness:} Current benchmarks primarily assess foundational video rendering in simplistic scenes, largely neglecting higher-order abilities such as composition and reasoning. Although benchmarks like OpenS2V~\citep{yuan2025opens2v} and VACE-Bench~\citep{jiang2025vace} include test cases for multimodal composition, they are insufficient in scale and completely omit reasoning evaluations.
Furthermore, most benchmarks rely on small, specialized tool models for assessment, unable to measure whether the generated videos truly align with user intentions in complex scenarios.
In contrast, designed with both complexity and comprehensiveness in mind, our IntelligentVBench encompasses diverse tasks, supports free-form inputs across multiple modalities, explicitly evaluates reasoning and compositional skills, and leverages a VLM-as-a-Judge~\citep{zheng2023judging} paradigm to ensure a robust evaluation.

\section{Training Data}

While conventional text-video paired data provides useful supervision, it falls short in supporting complex in-context reasoning and composition that involves interleaved text, images, and video inputs. 
Models trained exclusively on such data often struggle to capture nuanced semantic relationships across modalities.
To address these limitations, we incorporate large-scale vision-text interleaved data into our training corpus to enable richer multimodal interactions, utilizing videos sourced from both real-world and synthetic domains. In this section, we detail the training data sources, training tasks, and the data construction process in Sections~\ref{sec:train_data_source}, \ref{sec:train_task}, and~\ref{sec:train_data_cons}, respectively.

\subsection{Training Data Source}
\label{sec:train_data_source}

To construct a robust training corpus, we curate data from two complementary sources based on video provenance: real-world and synthetic domains. Real-world data encompasses a broad spectrum of visual content essential for capturing rich appearances, naturalistic motions, and complex scene dynamics; crucially, this anchors the generated videos to natural distributions and mitigates noticeable generative artifacts. However, because naturally occurring paired videos for highly conditioned tasks, such as video editing, are frequently sparse or inherently noisy, we incorporate synthetic data by leveraging off-the-shelf generation models~\citep{wu2025hunyuanvideo, wan2025wan, veo3} to rapidly synthesize target videos aligned with specific input conditions. While relying exclusively on synthetic data tends to introduce pronounced artificial biases, combining these two domains creates a synergistic effect that perfectly balances natural realism with task-specific conditioning density.

\subsection{Training Tasks}
\label{sec:train_task}

To ensure our training tasks facilitate richer multimodal interactions, we establish two core design principles: comprehensive coverage of \textit{\textbf{diverse multimodal scenarios}} and the systematic optimization of \textbf{\textit{hierarchical model capabilities}}. Accordingly, we structure our training framework around three primary competencies, each encompassing a diverse array of task formats.

\textbf{Foundational Video Generation Tasks:} This category integrates some foundational generation and editing tasks across three primary domains:
\textbf{(a)} Text-to-image and text-to-video synthesis with text-video or text-image paired data; 
\textbf{(b)} Instruction-guided video-to-video editing for both local and global modifications, such as background replacement, style transfer, object manipulation (addition, removal, or replacement), and text rendering; 
and \textbf{(c)} Key-frame(s)-to-video generation, which synthesizes continuous temporal sequences either from a single initial frame or via interpolation across multiple key-frames.

\textbf{Multimodal Composition Tasks.} Multimodal composition requires extracting and integrating distinct subjects or scenes from diverse inputs to synthesize a coherent video without unnatural blending artifacts. We formulate two primary tasks: \textbf{(a)} Interleaved Text-and-Multi-Image-to-Video generation, where the inputs contain multiple reference images (capturing key visual elements such as subjects or scenes) interleaved with text, requiring the model to accurately compose these elements into a cohesive video sequence; and \textbf{(b)} Text-Image-Video-to-Video generation, where the inputs consist of three modalities (image, text, and video), requiring the model to seamlessly integrate target visual elements extracted from reference images into the temporal dynamics of a source video.

\textbf{Reasoning-Augmented Tasks:} When user inputs are ambiguous, reasoning is essential to decipher the intended video content. 
Accordingly, we construct a reasoning-augmented dataset encompassing three main tasks: 
\textbf{(a)} Text-to-Video generation, where the model is trained to deduce comprehensive descriptions from brief, ambiguous input text queries prior to synthesis; 
\textbf{(b)} Intent-Driven Image-to-Video generation, where the model learns to formulate a reasoning trace detailing the temporal progression when visual and textual inputs lack explicit linkage (\textit{e.g.}, the text outlines abstract intents);
and \textbf{(c)} Event-Deductive Multi-Image-to-Video generation, where, given several highly disparate reference images as the key-frames, the model is optimized to bridge disparate reference key-frames by first uncovering implicit temporal dynamics via providing transition descriptions for these key-frames, before generating temporally coherent videos.

\subsection{Training Data Construction}
\label{sec:train_data_cons}

To construct our training tasks, we employ a dual-pipeline data construction strategy: output-first and input-first. 
In the output-first pipeline, we curate a diverse array of real-world videos from sources such as YouTube, cinematic clips, live-stream excerpts, and social media platforms to serve as ground-truth target videos.
Subsequently, an ensemble of auxiliary models is utilized to extract corresponding images or generate descriptive texts that act as task-specific inputs.
Conversely, the input-first pipeline begins by formulating the input conditions, leveraging video generation models, augmented by various tool models, to synthesize the corresponding ground-truth videos.
To facilitate both pipelines, we integrate a robust suite of models~\citep{yang2025qwen3, bai2025qwen3, carion2025sam, flux2, huang2026va, yang2026arcflow, leng2026preference, tu2025stableavatar, tu2025stableanimator, tu2024motioneditor, wu2025hunyuanvideo, wan2025wan, veo3}, such as Qwen3, Qwen3-VL, SAM3, FLUX2, and various video generation models, with Qwen3-VL additionally serving as an evaluator for rigorous data quality filtering. Next, we will provide a detailed exposition of the training data construction pipeline for each task.

\paragraph{\textbf{Foundational Video Generation Tasks:}}
Our primary training corpus for Text-to-Image (\textbf{T2I}), Text-to-Video (\textbf{T2V}), and Key-Frame(s)-to-Video generation (\textbf{I2V}) tasks is predominantly derived from extensive in-house datasets utilizing an output-first pipeline.
Specifically, the videos are mainly collected from web-sourced clips, and we leverage Qwen3-VL-235B to generate high-quality textual annotations as the user input.
To ensure the annotations align with specific task requirements, we design tailored prompting strategies: T2I and T2V prompts focus on visual semantic descriptions, whereas I2V prompts emphasize the dynamic transitions originating from the initial frame or the characterization of spatio-temporal offsets across multiple key-frames. 
Beyond this output-first paradigm, we also integrate an input-first strategy to produce a set of synthetic video data. Specifically, we first construct a carefully curated set of textual prompts and key-frames, and subsequently query Veo3~\citep{veo3} for high-quality video synthesis.

Furthermore, the training task for Instruction-guided video-to-video editing (\textbf{V2V}) encompasses both global and local modifications. Global edits primarily focus on background transformations and style changes, while local edits include fine-grained operations such as object addition, removal, replacement, and text manipulation within the video.
To construct a robust training corpus, we aggregate data from existing datasets, specifically OpenVE-3M~\citep{he2025openve} and Ditto~\citep{bai2025scaling}, and synthesize additional samples following their established pipelines. 
Finally, to guarantee high dataset fidelity, the entirety of this collected corpus undergoes rigorous quality filtration via Qwen3-VL-235B, ensuring the elimination of unsuccessful or low-quality edits.

\begin{figure}[t]
    \centering
    \includegraphics[width=\textwidth]{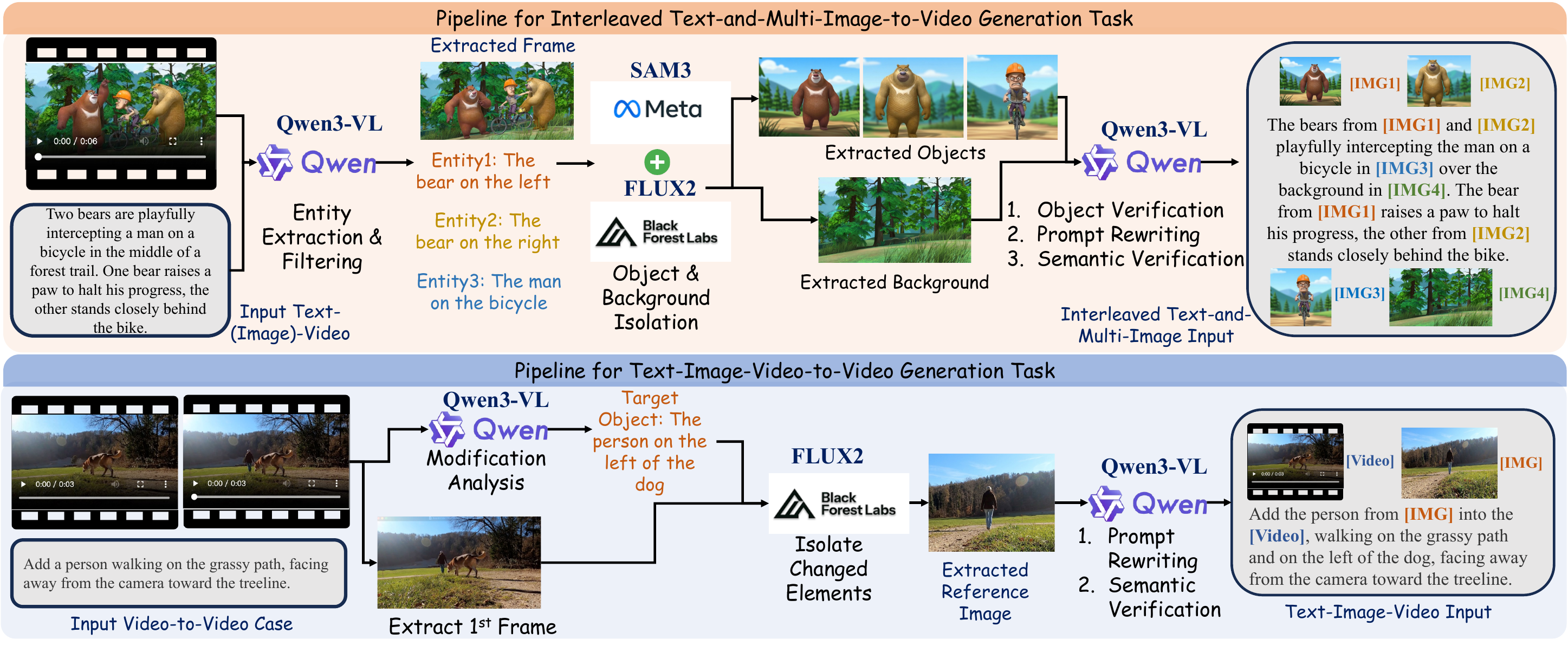} 
    \caption{Training data construction pipeline for Multimodal Composition Tasks.}
    \label{fig:pipeline}
\end{figure}

Notably, we observe that existing pipelines for local object addition V2V frequently yield physical inconsistencies, where newly introduced objects appear detached from the scene's underlying geometry and lighting. To rectify this, we invert the local object removal process, treating the post-removal video as the source input and the original, unedited video as the ground-truth target. 
By adapting the corresponding instructions from ``removal'' to ``addition'', we successfully generate high-quality V2V training samples for local object addition, characterized by physical realism and seamless environmental integration.

\paragraph{\textbf{Multimodal Composition Tasks.}}

The Multimodal Composition Tasks encompass two primary sub-tasks: Interleaved Text-and-Multi-Image-to-Video generation and Text-Image-Video-to-Video generation. The respective data construction pipelines are detailed below, also shown in Figure~\ref{fig:pipeline}.

Specifically, to facilitate \textbf{Interleaved Text-and-Multi-Image-to-Video generation}, we construct our dataset by systematically processing our image-to-video (I2V) task data.
Given a detailed video caption and its corresponding key-frame, we employ Qwen3-VL-235B to identify moving objects, representing them as named entities enriched with precise descriptive modifiers to ensure accurate localization. 
Qwen3-VL also acts as a filter to discard entities that feature ambiguous descriptions, isolated body parts, or objects absent from the key-frame.
Subsequently, for each validated entity, we utilize SAM3 to estimate its spatial location and temporal presence across the video sequence, followed by the application of FLUX2 to extract the corresponding object. 
To ensure appearance diversity and avoid pose duplication with the first frame, we leverage FLUX2 to extract these entity images from subsequent frames. 
Moreover, recognizing the potential extraction inaccuracies of FLUX2, we formulate four distinct prompt variations for each target object, leveraging Qwen3-VL as a verifier to confirm the identity alignment between the subject in the extracted image and the specified entity within the key-frame.
In addition to entity extraction, we also isolate the background from the first frame with FLUX2.
We then instruct Qwen3-VL to synthesize a comprehensive, reorganized prompt that integrates the extracted objects and the background, thereby formulating the interleaved text-and-multi-image input.
Ultimately, following a rigorous final verification by Qwen3-VL to ensure semantic consistency between the constructed interleaved input and the ground-truth video, we construct a large-scale, high-quality dataset tailored for Interleaved Text-and-Multi-Image-to-Video generation.

In contrast, we formulate the \textbf{Text-Image-Video-to-Video generation} task by repurposing existing video-to-video editing datasets. 
Given a source and target video pair, we first employ Qwen3-VL-235B, in conjunction with the original editing instruction, to analyze the specific modifications made in the target video, generating precise descriptive terms for the altered elements, such as a specific localized object or the overall video background.
We then extract the first frame of the target video and apply FLUX2 to isolate the specific visual elements that have changed relative to the source video to serve as our reference images.
Finally, we prompt Qwen3-VL a second time to rewrite the editing instructions and verify the semantic alignment between the input and output, thereby ensuring that the target elements detailed in the prompt are explicitly grounded in the extracted reference images.

\paragraph{\textbf{Reasoning-Augmented Tasks.}} The Reasoning-Augmented Tasks comprise three primary sub-tasks: Text-to-Video generation, Intent-Driven Image-to-Video generation, and Event-Deductive Multi-Image-to-Video generation. In addition to standard user inputs and ground-truth videos, each task incorporates a reasoning trace that explicitly bridges the input to the corresponding output. Next, we detail the pipelines for constructing the training data across these three tasks.

For \textbf{Text-to-Video generation}, we initialize our pipeline with a collection of brief, ambiguous text queries, where a subset of these queries has already been accompanied by ground-truth videos. 
To process the queries without corresponding videos, we leverage Qwen3-30B to generate detailed prompts acting as query-guidance.
These detailed prompts are subsequently fed into HunyuanVideo-1.5~\citep{wu2025hunyuanvideo} to generate the corresponding target videos.
Furthermore, for queries that already possess matching videos, we employ Qwen3-VL-235B to deeply analyze the video content and expand the initial brief query into a comprehensive, detailed prompt.
Ultimately, this dual-pathway approach yields robust training triplets, comprising the initial query, the detailed prompt guidance, and the associated video.
Therefore, it enables video generation models to ground video generation in sophisticated language-based reasoning.

To advance \textbf{Intent-Driven Image-to-Video generation}, we curate a dataset of cinematic text-video pairs wherein the text describes the action intent or behavioral premise driving the corresponding video scene.
For each sample, we leverage Qwen3-VL-235B to synthesize a detailed description of the temporal motion unfolding from the initial frame.
To ensure rigorous data quality, we further employ Qwen3-VL to strictly filter out instances where the generated motion lacks causal alignment with the driving intent.
On this basis, the action intent or behavioral premise serves as the user query, and the generated motion description functions as an intermediate reasoning trace.
It effectively enables the model to infer the reasoning trace prior to generating temporally coherent videos conditioned on both the first frame and the user query.

To facilitate \textbf{Event-Deductive Multi-Image-to-Video generation}, we process our collected video dataset by extracting key-frames and rigorously filtering them to ensure significant visual distinctions among key-frames.
Then we leverage Qwen3-VL-235B to generate two distinct types of textual annotations: a concise, single-sentence prompt that outlines the underlying intent or premise of the video, and a detailed prompt that intricately describes the dynamic temporal transitions across the given key-frames.
In formulating the training data, each instance incorporates the ordered key-frames and the corresponding ground-truth video, alongside a user query that either utilizes the concise prompt or relies solely on a generic instruction devoid of any explicit event cues, \textit{i.e.}, ``\textit{generate a complete video based on the provided key-frames}''.
Crucially, the detailed prompt acts as an explicit reasoning trace, thereby conditioning the model to logically deduce the underlying event trajectory implied by the key-frames before synthesizing the final video.
\section{Model: OmniWeaving}

\begin{figure}[t]
    \centering
    \includegraphics[width=\textwidth]{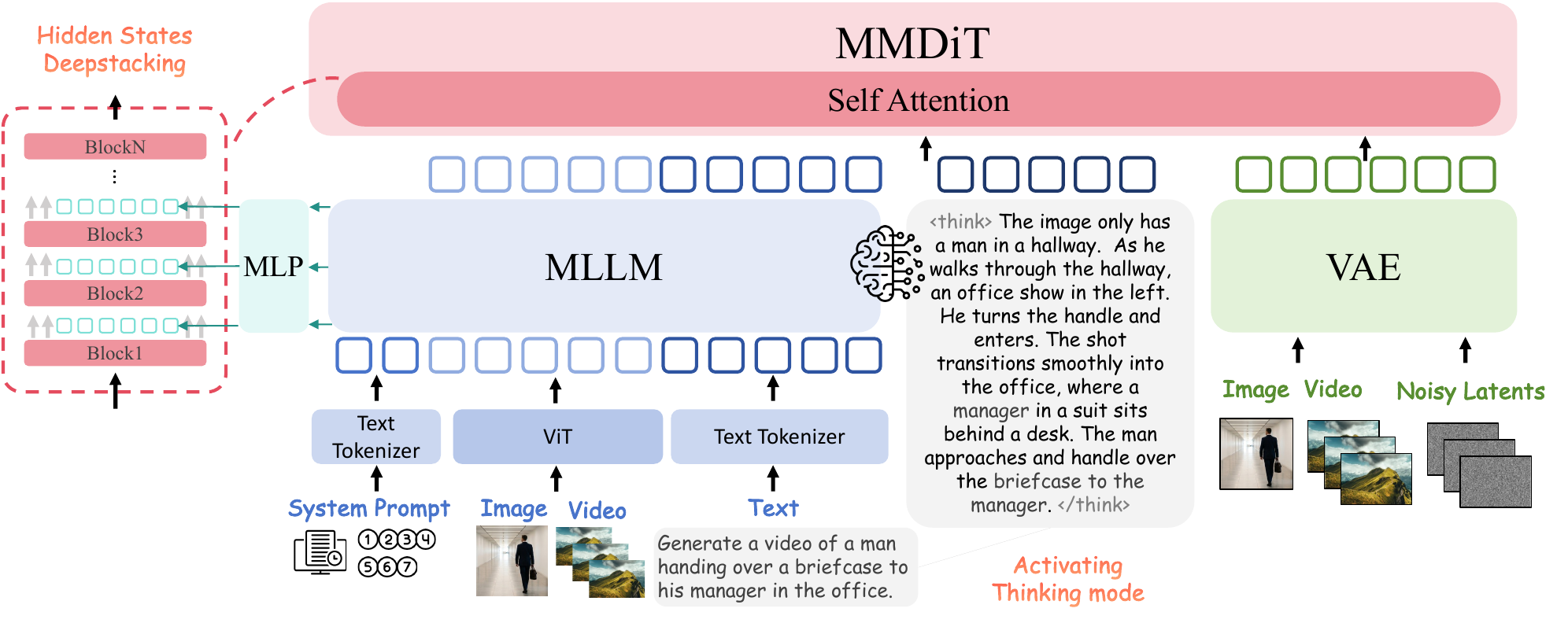} 
    \caption{OmniWeaving consists of an MLLM for multimodal understanding and an MMDiT for generation. On this basis, we activate the thinking mode of the MLLM and further introduce the DeepStacking mechanism.}
    \label{fig:arch}
\end{figure}

In this section, we introduce OmniWeaving, a unified framework for video generation and editing conditioned on free-form text, image, and video inputs. Crucially, the model possesses reasoning and compositional capabilities that are essential for intelligent unified video generation. We detail the model architecture, training strategy, and implementation details in Sections~\ref{sec:4.1}, \ref{sec:4.2}, and~\ref{sec:4.3}, respectively. More details are given in Appendix~\ref{app:1}.

\subsection{Model Architecture}
\label{sec:4.1}

Recognizing that robust visual comprehension is fundamental to unified free-form video generation, OmniWeaving is designed as an integrated architecture for multimodal understanding and generation.
As shown in Figure~\ref{fig:arch}, it comprises three core components: a Multimodal Large Language Model (MLLM), a Multimodal Diffusion Transformer (MMDiT), and a Variational Autoencoder (VAE).

First, \textbf{the MLLM} $\mathcal{M}$, with $L$ layers, acts as the core semantic parser. It projects a free-form multimodal input $\mathcal{X}$ into a high-level semantic space, yielding the layer-$l$ hidden states $\mathbf{H}^{(l)}=\mathcal{M}^{(l)}(\mathcal{X})$, and routes the last-layer hidden states $\mathbf{H}^{(L)}$ to the MMDiT via an MLP connector $\phi_0$.
Second, \textbf{the VAE} serves as a visual tokenizer, compressing visual inputs into low-level decodable latents $\mathbf{z}$ to provide fine-grained reconstruction signals. 
Finally, the \textbf{MMDiT} operates as the backbone diffusion model: its conditioning branch encodes the projected MLLM semantics $\phi_0(\mathbf{H}^{(L)})$, while the generative branch integrates the VAE latents $\mathbf{z}$ with noise, ultimately generating semantically aligned, high-fidelity videos. On this basis, we further introduce two extra improvements tailored for advanced reasoning and composition.

\textbf{(1) Activating Thinking Mode of the MLLM:}
Direct MLLM encoding of interleaved visual-text inputs often yields semantic ambiguity due to weak intra-correlations and unclear video creation intents. 
To address this issue, we elevate the MLLM from a passive feature extractor to an active reasoner. 
Specifically, given $\mathcal{X}$ and a reasoning-guidance prompt $p$, we activate the thinking mode of the MLLM to autoregressively generate a reasoning trace $\mathbf{r}\sim\mathcal{M}(\cdot\mid\mathcal{X},p)$, through which it autonomously deduces a semantically precise, enhanced prompt that explicitly specifies the target video. The last-layer hidden states of the concatenated sequence $[\mathcal{X};\mathbf{r}]$, which encode this enhanced prompt alongside the original input, are then projected and fed to the MMDiT conditioning branch as $\phi_0(\mathcal{M}^{(L)}([\mathcal{X};\mathbf{r}]))$, effectively bridging the cognitive gap between abstract user intent and pixel-level generation.

\textbf{(2) Hidden States DeepStacking:}
Compositional video generation involving multiple subjects or intricate scenes often relies on both low- and high-level semantic representations, whereas the last-layer states $\mathbf{H}^{(L)}$ alone are biased toward high-level abstraction.
Therefore, as shown in Figure~\ref{fig:arch}, we draw inspiration from the DeepStacking mechanism in Qwen3-VL~\citep{bai2025qwen3}, additionally extracting hidden states $\{\mathbf{H}^{(l_1)},\dots,\mathbf{H}^{(l_K)}\}$ from a broader range of intermediate MLLM layers to capture a rich semantic spectrum spanning from fine-grained details to high-level abstractions.
These multi-level features are projected into the MMDiT embedding space by layer-specific MLP connectors $\phi_k$ and then directly added to the hidden states $\mathbf{C}^{(k)}$ of the $k$-th layer of the MMDiT conditioning branch, \textit{i.e.}, $\mathbf{C}^{(k)} \leftarrow \mathbf{C}^{(k)} + \phi_k(\mathbf{H}^{(l_k)})$ for the first $K$ layers ($k=1,\dots,K$; $K=3$ in our implementation), effectively injecting multi-granular semantic guidance into the generative process.

\subsection{Training Strategy}
\label{sec:4.2}

Leveraging the proposed architecture and training data, we formulate a progressive, three-stage training paradigm, elevating OmniWeaving from a passive pixel renderer to an active and intelligent generalist.

\textbf{Stage 1: Modality Alignment Training.} As our generative backbone, \textit{i.e.}, HunyuanVideo-1.5~\citep{wu2025hunyuanvideo}, is highly proficient as a text-to-video expert but lacks prior exposure to multimodal hidden states produced by our MLLM, the primary objective of this initial stage is modality alignment between MLLM and MMDiT. 
We restrict the training to the most fundamental tasks: Text-to-Image (T2I), Text-to-Video (T2V), and Image-to-Video (I2V) from a single first frame. We keep the MLLM parameters strictly frozen and exclusively finetune the MMDiT and the MLP connectors with high-quality samples of these three tasks.
After this stage, OmniWeaving achieves performance comparable to the MMDiT backbone that uses its own text encoder.

\textbf{Stage 2: Multi-Task Free-Form Pretraining.} Having aligned the foundational modalities, we scale the training to encompass complex, heterogeneous inputs. Specifically, this stage incorporates all tasks detailed in Section~\ref{sec:train_task}, excluding the Reasoning-Augmented Tasks. 
However, we observe that introducing training tasks involving video input, such as video editing, at the very onset of the training impedes the model's learning efficiency on tasks that require processing interleaved image-text inputs.
Thus, we further partition this training phase into two distinct sub-stages: we initially exclude tasks involving video inputs, later integrating them for joint mixed training in the second sub-stage. 
Throughout this training stage, the MLLM parameters still remain frozen, and we exclusively finetune the MMDiT and the MLP connectors. 
Ultimately, we empower OmniWeaving to unify diverse video generation and editing tasks via multimodal instructions, establishing it as a versatile generalist framework with robust free-form compositional capabilities.

\textbf{Stage 3: Reasoning-Augmented Fine-Tuning.} In the final stage, we further elevate OmniWeaving's cognitive capabilities by introducing the reasoning-augmented tasks, jointly co-training alongside a curated high-quality subset from Stage 2. 
Crucially, we unfreeze the MLLM in this phase for end-to-end optimization alongside the MMDiT. 
Beyond the standard diffusion loss, the reasoning traces within the augmented data also introduce a next-token-prediction loss specifically designed to enhance the MLLM's reasoning proficiency. 
This establishes a ``comprehend-then-generate'' paradigm. 
Specifically, when confronted with ambiguous inputs, the MLLM learns to activate a ``thinking mode'' to extract explicit requirements via visual comprehension, thereby better guiding the MMDiT to synthesize highly aligned videos. 
Consequently, OmniWeaving evolves into an integrated, proactive reasoner and generator, functioning as an intelligent agent adept at managing diverse video generation tasks.

\begin{table}[t]
\centering
\caption{Training recipe of OmniWeaving.}
\vspace{-0.5em}
\label{tab:hyperparameters}
\resizebox{0.8\linewidth}{!}{
\begin{tabular}{l | c c c c}
\toprule
 & \multirow{2}{*}{\textbf{Stage1}} & \multirow{2}{*}{\textbf{Stage2}} & \multicolumn{2}{c}{\textbf{Stage3}} \\ \cline{4-5}
 & & & \textbf{MLLM} & \textbf{MMDiT} \\
\hline
\textbf{Hyperparameters} & & & & \\
Learning rate & $2.0 \times 10^{-5}$ & $2.0 \times 10^{-5}$ & $3.0 \times 10^{-5}$ & $1.0 \times 10^{-5}$ \\
LR scheduler & Constant & Constant & Cosine & Constant \\
Min Learning rate & $2.0 \times 10^{-5}$ & $2.0 \times 10^{-5}$ & $2.0 \times 10^{-6}$ & $1.0 \times 10^{-5}$ \\
Weight decay & 0.01 & 0.01 & 0.05 & 0.01 \\
Gradient norm clip & 1.0 & 1.0 & 1.0 & 1.0 \\

Optimizer &  Muon & Muon & AdamW & Muon \\
GPUs &  $128 \times$ H20s &  $400 \times$ H20s &  \multicolumn{2}{c}{$400 \times$ H20s} \\
Batch Size & 128 & 400 & \multicolumn{2}{c}{400} \\
Loss weight (CE : MSE) & - & - & \multicolumn{2}{c}{$0.25 : 1$} \\
Warm-up steps & 250 & 1200 & \multicolumn{2}{c}{250} \\
Training steps & 5K & 50K & \multicolumn{2}{c}{3K} \\
Resolution & 480p & 480p & \multicolumn{2}{c}{480p} \\
Text dropout ratio & 0.1 & 0.1 & \multicolumn{2}{c}{0.1} \\

Diffusion timestep shift & \multicolumn{4}{c}{1.0 for image and 4.0 for video} \\
\hline
\textbf{Data sampling ratio} & & & \multicolumn{2}{c}{} \\
Text-to-Image & 0.10 & 0.05 $\to$ 0.03 & \multicolumn{2}{c}{0.00} \\
Text-to-Video & 0.40 & 0.15 $\to$ 0.07 & \multicolumn{2}{c}{0.05} \\
First-Frame-to-Video & 0.50 & 0.15 $\to$ 0.06 & \multicolumn{2}{c}{0.05} \\
Key-Frames-to-Video & 0.00 & 0.25 $\to$ 0.14 & \multicolumn{2}{c}{0.05} \\
Video-to-Video & 0.00 & 0.00 $\to$ 0.25 & \multicolumn{2}{c}{0.15} \\
Interleaved Text-and-Multi-Image-to-Video & 0.00 & 0.40 $\to$ 0.20 & \multicolumn{2}{c}{0.15} \\
Text-Image-Video-to-Video & 0.00 & 0.00 $\to$ 0.25 & \multicolumn{2}{c}{0.15} \\
Reasoning-Augmented Tasks & 0.00 & 0.00 & \multicolumn{2}{c}{0.40} \\
\bottomrule
\end{tabular}
}
\end{table}

\subsection{Implementation Details}
\label{sec:4.3}
We implement OmniWeaving by integrating Qwen2.5-VL~\citep{bai2025qwen25vltechnicalreport} as the foundational MLLM and HunyuanVideo-1.5~\citep{wu2025hunyuanvideo} as the generative MMDiT. To instantiate our DeepStacking mechanism, we extract the hidden states from the 8th, 16th, and 24th layers of the MLLM. Each of these states is projected through its own trainable MLP connector ($\phi_1$--$\phi_3$ in Section~\ref{sec:4.1}) and subsequently added directly to the corresponding hidden states of the first three conditioning layers within the MMDiT.
To mitigate the computational overhead of attention in the MMDiT, we adopt the sparse attention implementation from SSTA (Selective and Sliding Tile Attention)~\citep{wu2025hunyuanvideo}. Furthermore, we employ the Muon optimizer~\citep{liu2025muon} to accelerate training convergence.
The training process is structured into three distinct stages, spanning 5k, 50k, and 3k steps, with learning rates set to 2e-5, 2e-5, and 1e-5, respectively. During the second (multi-task free-form pretraining) stage, we initially exclude tasks involving video inputs for the first 20k steps. In the subsequent 30k steps, we incorporate these tasks involving video inputs, encompassing all tasks detailed in Section~\ref{sec:train_task}, with the  exception of the Reasoning-Augmented Tasks. These Reasoning-Augmented Tasks are exclusively introduced during the third training stage (Reasoning-Augmented Fine-Tuning).
To facilitate large-scale optimization, we train our model with 400 NVIDIA H20 GPUs, maintaining a global batch size of 400. Our primary focus is on generating videos at a 480p resolution; accordingly, all training videos contain between 33 and 161 frames, with aspect ratios strictly bounded between 0.25 and 4.0. The training hyperparameters and data sampling ratio for each stage are summarized in Table~\ref{tab:hyperparameters}.

\section{Benchmark: IntelligentVBench}

\begin{figure}[htbp]
    \centering
    \includegraphics[width=\textwidth]{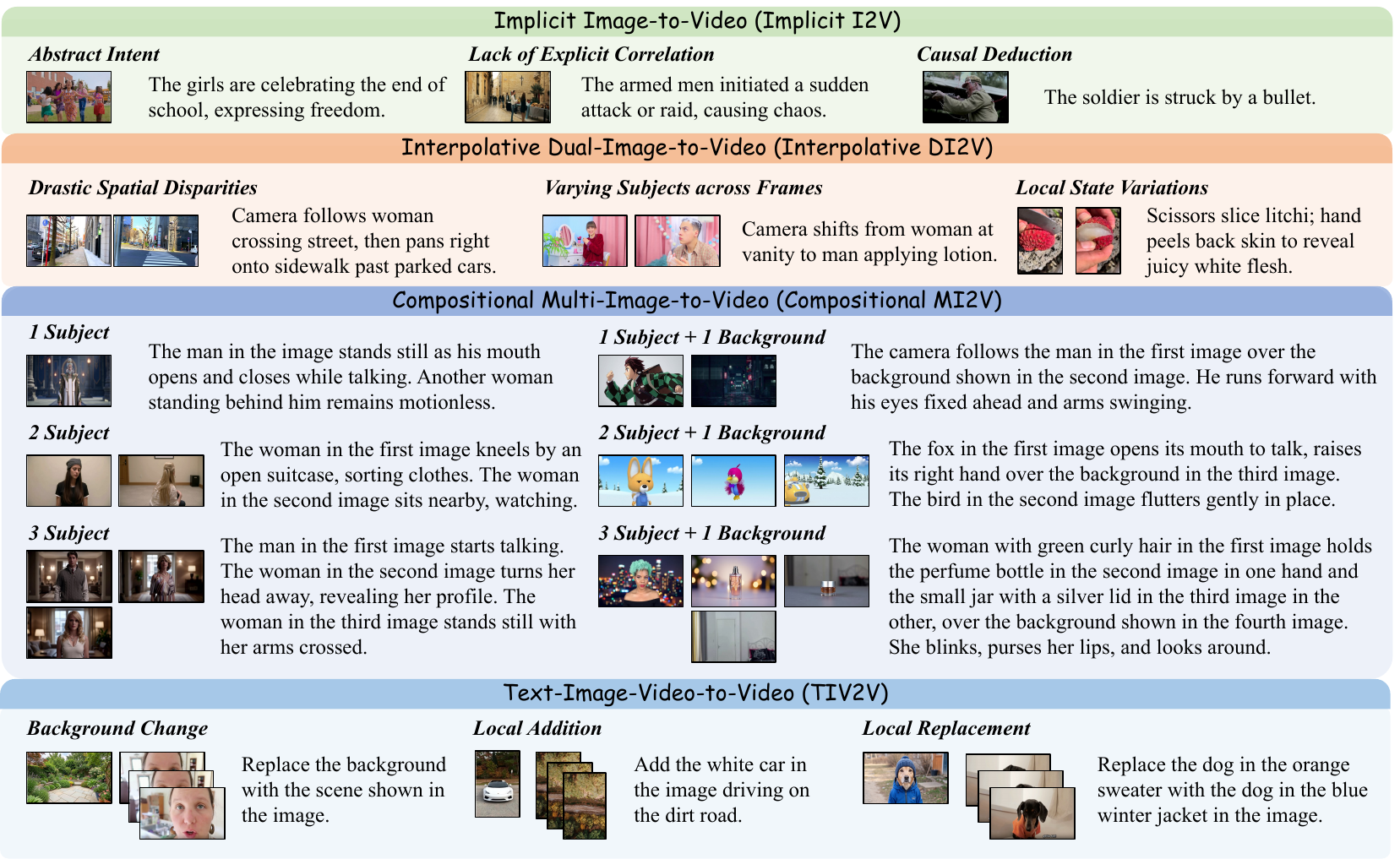} 
    \caption{Examples of each task type in IntelligentVBench.}
    \label{fig:bench}
\end{figure}

As video generation models rapidly advance, traditional benchmarks struggle to evaluate their true capabilities. As discussed in Section~\ref{sec:2}, they often lack complexity by confining evaluations to isolated, single-shot tasks, and lack comprehensiveness by prioritizing basic rendering in simplistic scenes over the crucial roles of multi-element composition and abstract reasoning.
To bridge this gap, we introduce IntelligentVBench, a novel evaluation suite designed to benchmark next-level intelligent video generation by integrating abstract reasoning and free-form composition across a diverse array of complex tasks. A comparison between IntelligentVBench and existing video generation benchmarks is shown in Table~\ref{tab:bench}.
In Figure~\ref{fig:bench}, we further present examples of each task type in IntelligentVBench.

\begin{table*}[t!]
\centering 

\caption{\label{tab:bench}Compare IntelligentVBench with existing video generation benchmarks.}
\vspace{-0.5em}
\resizebox{1.0\textwidth}{!}{
\begin{tabular}{l|ccc|ccc|ccc|c}
\toprule
\multirow{2}{*}{\textbf{Benchmark}}  & \multirow{2}{*}{\textbf{\#Size}} & \textbf{Multi-} & \textbf{Interleaved}
& \multicolumn{3}{c|}{\textbf{Input modality}} & \multicolumn{3}{c|}{\textbf{Ability to Measure}} &   \multirow{2}{*}{\textbf{VLM-as-a-Judge}} \\ 
          &        &    \textbf{Task}   &    \textbf{Input}      & Image & Multi-Images & Video & Foundation & Composition & Reasoning &   \\ \hline

VBench \citep{huang2024vbench} & 946 & \xmark & \xmark & \xmark & \xmark & \xmark & \cmark & \xmark & \xmark & \xmark \\
VBench++ \citep{huang2025vbench++} & 2384 & \cmark & \xmark & \cmark & \xmark & \xmark & \cmark & \xmark & \xmark & \xmark \\
OpenVE-Bench \citep{he2025openve} & 431 & \xmark & \xmark & \xmark & \xmark & \cmark & \cmark & \cmark & \xmark & \cmark \\
TGVE+ \citep{singer2024video} & 1417 & \xmark & \xmark & \xmark & \xmark & \cmark & \cmark & \cmark & \xmark & \xmark \\
OpenS2V-Eval \citep{yuan2025opens2v} & 180 & \xmark & \cmark & \cmark & \cmark & \xmark & \cmark & \cmark & \xmark & \xmark  \\
VACE-Bench \citep{jiang2025vace} & 480 & \cmark & \cmark & \cmark & \cmark & \cmark & \cmark & \cmark & \xmark & \xmark \\
\hline
\textbf{IntelligentVBench} & 1030 & \cmark & \cmark & \cmark & \cmark & \cmark & \cmark & \cmark & \cmark & \cmark \\

 \bottomrule
\end{tabular}%
}
\end{table*}

\subsection{Design Principles}
To comprehensively evaluate the complex capabilities required of unified video generation models, the design of IntelligentVBench is guided by two core principles, closely mirroring the demands of real-world human queries.

\textbf{Heterogeneous and Free-Form Inputs}. 
To better reflect the complexities of real-world human queries, our benchmark aims to employ test cases featuring free-form, interleaved vision-text inputs. 
Specifically, inputs should consist of diverse combinations of multiple images, text, and, when necessary, reference video contexts. 
Crucially, images within the prompt serve multifaceted roles: 
Some act as spatial anchors, providing specific objects intended for scene composition, while others serve as temporal anchors to delineate event progression. 
This design compels models to dynamically interpret unstructured, multimodal conditions rather than overfitting to rigid input templates.

\textbf{Evaluating Composition and Reasoning:}
We consider composition and reasoning to be the two most critical capabilities for advanced intelligent video generation, thus structuring our evaluation around them.
Specifically, for \textbf{(a) Composition}, given complex inputs comprising diverse subjects and scenes, we require the model to not only achieve accurate static spatial composition, but also to temporally synthesize the dynamic interactions among these entities and their interplay with contextual background elements throughout the generated frames. Furthermore, regarding \textbf{(b) Reasoning}, many cases within our benchmark cannot be resolved through direct explicit scene rendering; instead, the model may first need to deduce temporal correlations across significantly disparate reference images, interpret textual descriptions of object behaviors that remain absent from the visual inputs, or process prompts that merely articulate the abstract intent underlying an event.

\subsection{Task Taxonomy}
\label{sec:5.2}

Guided by the design principles, we construct four distinct tasks within IntelligentVBench: Implicit Image-to-Video (Implicit I2V), Interpolative Dual-Image-to-Video (Interpolative DI2V), Compositional Multi-Image-to-Video (Compositional MI2V), and Text-Image-Video-to-Video (TIV2V). The first two are explicitly designed to evaluate reasoning, whereas the latter two focus on composition.

\textbf{Implicit I2V:} 
Although structurally similar to standard I2V tasks, where an image provides the initial frame and a text prompt guides the generation, this task emphasizes implicit reasoning. It requires the model to logically unfold temporal events under high ambiguity, \textit{e.g.}, when the prompt expresses abstract intent without concrete actions, or when the described behavior lacks explicit grounding in the input image, or when text provides only preliminary assumptions requiring causal deduction.

\textbf{Interpolative DI2V:} 
It requires synthesizing a temporally coherent video conditioned on two boundary frames and a text prompt, presenting three primary challenges: (a) large spatial disparities, requiring the generation of plausible physical trajectories or complex camera motions; (b) multi-subject dynamics, necessitating the inference of complex intermediate states for various characters; and (c) local environmental variations, which demand deducing the underlying events that drive these visual transformations.

\textbf{Compositional MI2V:} 
Given 1--4 reference images of subjects or backgrounds, alongside text prompts that may introduce additional objects, the objective is to seamlessly integrate these disparate elements into a coherent video while preserving the given visual identities.

\textbf{TIV2V:} This task assesses cross-modal compositional editing. Given a source video, reference images, and a text instruction, the model must extract specific concepts from the references and integrate them into the source video, while preserving the temporal consistency and visual fidelity of non-target elements.

In total, our benchmark comprises \textbf{1,030} test cases. Specifically, the \textbf{Implicit I2V} task contains 250 test cases; to construct highly challenging data, all input images are curated from movie clips featuring complex motions, with PhD experts formulating implicit instructions based on the corresponding cinematic plots. 
The \textbf{Interpolative DI2V} task also includes 250 test cases, where the paired input images are sourced from social networks or film segments to ensure they are contextually related yet exhibit significant visual variance. 
Furthermore, the \textbf{Compositional MI2V} task encompasses 320 test cases, each accepting one to four input images.
One to three images define the subject(s), and at most one image specifies the scene context. 
We further categorize this task into three subtasks based on the number of subject input images: 130 cases feature a single subject image input (with 78 cases of these including an additional scene image), 120 cases feature two subject image inputs (with 40 cases including an additional scene image), and 70 cases feature three subject image inputs (with 36 cases including an additional scene image).
Finally, the \textbf{TIV2V} task comprises a total of 210 test cases spanning three distinct operational objectives: inserting an object from a reference image into the source video (71 test cases), substituting a specific visual element within the video with one from the reference image (73 test cases), and replacing the video's original background with the scene depicted in the reference image (66 test cases).

\subsection{Evaluation Metrics}

To comprehensively assess model performance on IntelligentVBench, we adopt the ``VLM-as-a-Judge''~\citep{zheng2023judging} paradigm, employing Gemini2.5-Pro~\citep{comanici2025gemini} as an automatic evaluator to establish three essential metrics for each task:

\textbf{Instruction Following} assesses how well the video executes the semantic and logical intent of the text prompt. 
For example, for the Implicit I2V task, it evaluates whether the model successfully captures the semantic correlation between the input image and the prompt to generate videos that accurately align with the user's underlying intent. 
For the Interpolative DI2V task, it gauges the intuitive transition logic between key-frames. 
For the Compositional MI2V task, it examines whether the model accurately constructs all target visual elements with plausible interactions among multiple subjects following the prompt. 
Finally, for the TIV2V task, it measures the model's capacity to seamlessly integrate the provided video and image in strict adherence to the instruction.

\textbf{Condition Preserving} evaluates the fidelity with which a generated video reconstructs and retains the identity, structure, and intricate details of provided visual references. Specifically, in Implicit I2V, where a single input image serves as the first frame, this metric verifies whether the first frame of the generated video faithfully reproduces the input image, with subsequent frames remaining stylistically consistent with it. In Interpolative DI2V, it verifies whether both the first and last frames anchor flawlessly to the two conditioning images without mutational drift. For Compositional MI2V, it measures subject and scene consistency by assessing whether the generated output faithfully preserves the multiple distinct identities and spatial layouts dictated by the input references. Finally, for TIV2V, the metric serves a dual purpose: it gauges the preservation of the source video's unedited regions and original temporal dynamics, while simultaneously verifying that the newly introduced objects or scenes faithfully conform to the provided reference images.

\textbf{Overall Visual Quality} evaluates the general aesthetic quality, temporal consistency, motion smoothness, and physical naturalness of the generated video. This metric focuses on ensuring the outputs are fluid, obey real-world physics, and are devoid of noticeable AI artifacts or high-frequency flickering. 
Importantly, in the TIV2V task, if the source video inherently suffers from compromised aesthetic quality and the accompanying text prompt does not explicitly mandate aesthetic restoration, it should not be penalized during evaluation for retaining visual artifacts that were already present in the input video.

Each metric is rated on a 1–5 scale based on carefully crafted prompts to ensure precise scoring.  
Furthermore, we establish two metrics to evaluate overall performance. Let $\mathcal{IF}_i$, $\mathcal{CP}_i$, and $\mathcal{VQ}_i$ denote the scores of the $i$-th of the $N$ test cases in a task; then
\begin{equation*}
\mathbf{AVG}=\frac{1}{N}\sum_{i=1}^{N}\frac{\mathcal{IF}_i+\mathcal{CP}_i+\mathcal{VQ}_i}{3}, \qquad
\mathbf{MIN}=\frac{1}{N}\sum_{i=1}^{N}\min\left(\mathcal{IF}_i,\mathcal{CP}_i,\mathcal{VQ}_i\right).
\end{equation*}
$\mathbf{AVG}$ equals the mean of the three task-level metrics, whereas $\mathbf{MIN}$ takes the minimum within each case before averaging, so a case that fails on any single metric is penalized and $\mathbf{MIN}$ is never higher than the smallest of the three task-level metrics.
The evaluation prompt template for Implicit I2V task is detailed in Figure~\ref{fig:bench_i2v} in Appendix~\ref{app:3}.
The evaluation prompt template for Interpolative DI2V task is detailed in Figure~\ref{fig:bench_di2v}.
The evaluation prompt templates for Compositional MI2V task are detailed in Figure~\ref{fig:bench_mi2v_single} and Figure~\ref{fig:bench_mi2v}.
Furthermore, we design three distinct evaluation prompt templates tailored to the specific sub-tasks within the TIV2V task, as illustrated in Figure~\ref{fig:bench_tiv2v_add}, Figure~\ref{fig:bench_tiv2v_back}, and Figure~\ref{fig:bench_tiv2v_change}.

\section{Experiments}
\subsection{Experimental Setup}

All evaluations are conducted in a zero-shot setup. In our comprehensive evaluation across both the proposed IntelligentVBench and existing benchmarks, we categorize baseline models into two primary paradigms: unified video generation models and task-specific video generation models. 
As for unified video generation baselines, we introduce three academic models, VACE~\citep{jiang2025vace}, VINO~\citep{chen2026vino}, and UniVideo~\citep{wei2025univideo}, as well as the industrial open-weight model MiniMax-H3~\citep{minimax}.
Specifically, we evaluate two architectural variants of VACE~\citep{jiang2025vace}, distinguished by their underlying backbones: VACE-Wan2.1~\citep{wan2025wan} and VACE-LTX~\citep{hacohen2024ltx}.
Furthermore, both VINO~\citep{chen2026vino} and UniVideo~\citep{wei2025univideo} employ a unified architecture that couples an MLLM~\citep{bai2025qwen3} with a downstream diffusion model~\citep{kong2024hunyuanvideo}.
Building on this, we assess two variants of UniVideo: UniVideo(query), which utilizes a query token mechanism~\citep{pan2025transfer} to extract conditional embeddings from the MLLM, and UniVideo(hidden), which directly leverages the MLLM's final hidden states to condition the diffusion model.
Finally, MiniMax-H3 is a recently released industrial open-weight unified video generation model with 33B parameters, distributed under a custom community license without a technical report on its training data or recipe.
Unlike the above unified baselines, it does not rely on a single checkpoint: one checkpoint handles text-to-video and first/last-image(s)-to-video, and another checkpoint handles reference-to-video.

To establish task-specific baselines, we evaluate a diverse suite of models tailored to each task within IntelligentVBench.
For Implicit I2V, we include CogVideoX-I2V~\citep{yang2024cogvideox}, Wan2.1-I2V~\citep{wan2025wan}, HunyuanVideo-I2V~\citep{kong2024hunyuanvideo}, HunyuanVideo1.5-I2V~\citep{wu2025hunyuanvideo}, and Wan2.2-I2V~\citep{wan2025wan} as the baselines. 
For Interpolative DI2V, we employ Wan2.1-FLF2V~\citep{wan2025wan} as the baseline. 
And for Compositional MI2V, we utilize SkyReels-A2~\citep{fei2025skyreels}, SkyReels-V3~\citep{li2026skyreels}, MAGREF~\citep{deng2025magref}, and Phantom~\citep{liu2025phantom}.
Notably, no specialized models currently exist for the TIV2V task. 
Furthermore, for the video editing task on OpenVE-Bench, we incorporate six specialized editing models: OmniVideo~\citep{tan2025omni}, InsViE~\citep{wu2025insvie}, Lucy-Edit~\citep{team2025lucy}, ICVE~\citep{liao2025context}, Ditto~\citep{bai2025scaling}, and OpenVE-Edit~\citep{he2025openve}.
Finally, for the text-to-video generation task on VBench, we include four task-specific baselines, \textit{i.e.}, StepVideo~\citep{ma2025step}, CogVideoX~\citep{yang2024cogvideox}, HunyuanVideo~\citep{kong2024hunyuanvideo}, and Wan2.1~\citep{wan2025wan}.

\subsection{Main Results on IntelligentVBench}

\begin{table*}[t]
    \centering
    \caption{ \label{tab:main1} Main results of the Implicit I2V, Interpolative DI2V, and TIV2V tasks in IntelligentVBench. The best results are in \textbf{bold} for {\color[HTML]{F88825} specialized-} and {\color[HTML]{319B62} unified-}models, with the second best \underline{underlined}. \#GenParams reports only the generation parameters.} 
    \vspace{-0.5em}
    \resizebox{1.0\linewidth}{!}{
        \begin{tabular}{lc|ccccc|ccccc|ccccc}
            \toprule
            \multirow{2}{*}{Model}    & \multirow{2}{*}{\#GenParams}&\multicolumn{5}{c}{\textbf{Implicit I2V}}  &
                    \multicolumn{5}{c}{\textbf{Interpolative DI2V}}  &
                    \multicolumn{5}{c}{\textbf{TIV2V}}  \\
            &  & $\mathcal{IF}\uparrow$ & $\mathcal{CP}\uparrow$ & $\mathcal{VQ}\uparrow$ & $\mathbf{MIN}$ & $\mathbf{AVG}$ & $\mathcal{IF}\uparrow$ & $\mathcal{CP}\uparrow$ & $\mathcal{VQ}\uparrow$ & $\mathbf{MIN}$ & $\mathbf{AVG}$ & $\mathcal{IF}\uparrow$ & $\mathcal{CP}\uparrow$ & $\mathcal{VQ}\uparrow$ & $\mathbf{MIN}$ & $\mathbf{AVG}$ \\ \hline
            \multicolumn{17}{c}{\textit{Specialized Video Generation Models}} \\ \hline

            CogVideoX-I2V & 5B & 3.53  & \color[HTML]{F88825} \underline{3.74}  & 2.90  & 2.68  & 3.39 & - & - & - & - & - & - & - & - & - & -\\
            Wan2.1-I2V  & 14B & 3.76  & 3.33  & 3.34  & 2.78  & 3.48 & - & - & - & - & - & - & - & - & - & -   \\
            Wan2.1-FLF2V & 14B & - & - & - & - & - & 4.48 & 4.28 & 4.49 & 3.98 & 4.42 & - & - & - & - & -\\
            HunyuanVideo-I2V & 13B & 3.37  & \color[HTML]{F88825} \textbf{4.17}  & \color[HTML]{F88825} \textbf{3.87}  & 3.00  & \color[HTML]{F88825} \underline{3.80} & - & - & - & - & - & - & - & - & - & -  \\
            HunyuanVideo1.5-I2V & 8.3B & \color[HTML]{F88825} \underline{3.98}  & 3.44  & 3.67  & \color[HTML]{F88825} \underline{3.12}  & 3.70 & - & - & - & - & - & - & - & - & - & - \\
            Wan2.2-I2V  & 14B & \color[HTML]{F88825} \textbf{4.21}  & 3.70  & \color[HTML]{F88825} \underline{3.68}  & \color[HTML]{F88825} \textbf{3.30}  & \color[HTML]{F88825} \textbf{3.86} & - & - & - & - & - & - & - & - & - & -\\ 
            \hline  

            \multicolumn{17}{c}{\textit{Unified Video Generation Models}} \\ \hline
            MiniMax-H3 & 33B & \color[HTML]{319B62} \textbf{4.55} & \color[HTML]{319B62} \textbf{3.88} & \color[HTML]{319B62} \textbf{4.12} & \color[HTML]{319B62} \textbf{3.60} & \color[HTML]{319B62} \textbf{4.19} & \color[HTML]{319B62} \underline{4.71} & \color[HTML]{319B62} \underline{4.37} & \color[HTML]{319B62} \underline{4.51} & \color[HTML]{319B62} \textbf{4.11} & \color[HTML]{319B62} \underline{4.53} & \color[HTML]{319B62} \textbf{4.36} & \color[HTML]{319B62} \textbf{4.21} & \color[HTML]{319B62} \textbf{4.03} & \color[HTML]{319B62} \textbf{3.71} & \color[HTML]{319B62} \textbf{4.20}\\ \hline
            VACE-Wan2.1  &  14B & 3.66  & 3.22  & 3.32  & 2.76  & 3.40 & 4.01 & 3.61 & 3.96 & 3.32 & 3.86 & 1.46 & 1.42 & 1.71 & 1.27 & 1.53\\
            VACE-LTX  & 2B & 3.04  & 3.28  & 2.28  & 2.20  & 2.87 & 3.51 & 3.13 & 2.67 & 2.59 & 3.10 & 1.43 & 1.36 & 1.25 & 1.20 & 1.35\\
            VINO  & 13B & 3.70  & 2.64  & 2.96  & 2.28  & 3.10 & 2.38 & 1.73 & 3.29 & 1.65 & 2.47 & 2.86 & 2.90 & 2.52 & 2.11 & 2.76\\
            UniVideo(query)  & 13B & 3.66  & 3.37  & 3.16  & 2.77  & 3.39 & 2.31 & 1.85 & 3.27 & 1.76 & 2.48 & 3.22 & 3.91 & 3.26 & 2.66 & 3.46\\
            UniVideo(hidden)  & 13B & 3.74  & \color[HTML]{319B62} \underline{3.82}  & 3.28  & 2.95  & 3.61 & 2.36 & 1.98 & 3.40 & 1.84 & 2.58 & 3.13 & 4.01 & 2.93 & 2.56 & 3.36\\
            \hline
            \textbf{Ours (w/o think)} & 8.3B & 4.05  & 3.75  & 3.36  & 3.11  & 3.72 & 4.60 & 4.22 & \color[HTML]{319B62} \textbf{4.52} & \color[HTML]{319B62} \underline{3.99} & 4.45 & \color[HTML]{319B62} \underline{4.00} & \color[HTML]{319B62} \underline{4.04} & \color[HTML]{319B62} \underline{3.65} & \color[HTML]{319B62} \underline{3.31} & \color[HTML]{319B62} \underline{3.89}\\
            \textbf{Ours (think)} & 8.3B & \color[HTML]{319B62} \underline{4.33}  & \color[HTML]{319B62} \underline{3.82}  & \color[HTML]{319B62} \underline{3.63}  & \color[HTML]{319B62} \underline{3.34}  & \color[HTML]{319B62} \underline{3.93} & \color[HTML]{319B62} \textbf{4.74} & \color[HTML]{319B62} \textbf{4.38} & 4.50 & \color[HTML]{319B62} \textbf{4.11} & \color[HTML]{319B62} \textbf{4.54} & - & - & - & - & -\\

            \bottomrule
        \end{tabular}}
\end{table*}

\begin{table*}[t]
    \centering
    \caption{ \label{tab:main2} Main results of Compositional MI2V task in IntelligentVBench. We define 3 sub-categories based on the number of subjects presented in the 1–4 input images.}
    \vspace{-0.5em}
    \resizebox{1.0\linewidth}{!}{
        \begin{tabular}{lc|ccccc|ccccc|ccccc}
            \toprule
            \multirow{2}{*}{Model}    & \multirow{2}{*}{\#GenParams}&\multicolumn{5}{c}{\textbf{1 Subject (w/ or w/o BKG)}}  &
                    \multicolumn{5}{c}{\textbf{2 Subjects (w/ or w/o BKG)}}  &
                    \multicolumn{5}{c}{\textbf{3 Subjects (w/ or w/o BKG)}}  \\
            &  & $\mathcal{IF}\uparrow$ & $\mathcal{CP}\uparrow$ & $\mathcal{VQ}\uparrow$ & $\mathbf{MIN}$ & $\mathbf{AVG}$ & $\mathcal{IF}\uparrow$ & $\mathcal{CP}\uparrow$ & $\mathcal{VQ}\uparrow$ & $\mathbf{MIN}$ & $\mathbf{AVG}$ & $\mathcal{IF}\uparrow$ & $\mathcal{CP}\uparrow$ & $\mathcal{VQ}\uparrow$ & $\mathbf{MIN}$ & $\mathbf{AVG}$ \\ \hline
            \multicolumn{17}{c}{\textit{Specialized Video Generation Models}} \\ \hline

SkyReels-A2  & 14B & \color[HTML]{F88825} \textbf{3.51} & \color[HTML]{F88825} \textbf{4.08} & \color[HTML]{F88825} \underline{4.46} & \color[HTML]{F88825} \textbf{3.24} & \color[HTML]{F88825} \textbf{4.02} & \color[HTML]{F88825} \underline{3.22} & \color[HTML]{F88825} \underline{3.76} & 4.37 & \color[HTML]{F88825} \underline{2.97} & \color[HTML]{F88825} \underline{3.78} & 1.64 & 1.76 & 2.50 & 1.56 & 1.97 \\
SkyReels-V3  & 14B &  \color[HTML]{F88825} \underline{3.46} & \color[HTML]{F88825} \underline{3.71} & \color[HTML]{F88825} \textbf{4.65} & \color[HTML]{F88825} \underline{2.98} & \color[HTML]{F88825} \underline{3.94} & \color[HTML]{F88825} \textbf{3.28} & \color[HTML]{F88825} \textbf{3.84} & \color[HTML]{F88825} \textbf{4.44} & \color[HTML]{F88825} \textbf{3.04} & \color[HTML]{F88825} \textbf{3.86} & \color[HTML]{F88825} \textbf{2.59} & \color[HTML]{F88825} \textbf{3.10} & \color[HTML]{F88825} \underline{4.30} & \color[HTML]{F88825} \textbf{2.37} & \color[HTML]{F88825} \textbf{3.33} \\
MAGREF  & 14B & 3.15 & 2.48 & 4.32 & 2.18 & 3.32 & 3.04 & 2.81 & 4.33 & 2.44 & 3.39 & \color[HTML]{F88825} \underline{2.50} & 2.21 & \color[HTML]{F88825} \textbf{4.46} & 2.07 & 3.06 \\
Phantom  & 14B & 3.21 & 2.95 & 4.29 & 2.47 & 3.48 & 2.88 & 3.42 & \color[HTML]{F88825} \underline{4.38} & 2.62 & 3.56 & 2.36 & \color[HTML]{F88825} \underline{2.79} & 4.21 & \color[HTML]{F88825} \underline{2.20} & \color[HTML]{F88825} \underline{3.12} \\ \hline

\multicolumn{17}{c}{\textit{Unified Video Generation Models}} \\ \hline
MiniMax-H3 & 33B & \color[HTML]{319B62} \textbf{4.44} & \color[HTML]{319B62} \textbf{4.75} & \color[HTML]{319B62} \textbf{4.78} & \color[HTML]{319B62} \textbf{4.27} & \color[HTML]{319B62} \textbf{4.66} & \color[HTML]{319B62} \textbf{4.40} & \color[HTML]{319B62} \textbf{4.75} & \color[HTML]{319B62} \textbf{4.75} & \color[HTML]{319B62} \textbf{4.24} & \color[HTML]{319B62} \textbf{4.63} & \color[HTML]{319B62} \textbf{4.12} & \color[HTML]{319B62} \textbf{4.48} & \color[HTML]{319B62} \textbf{4.65} & \color[HTML]{319B62} \textbf{3.88} & \color[HTML]{319B62} \textbf{4.42} \\ \hline
VACE-Wan2.1  &  14B & 3.88 & 4.48 & \color[HTML]{319B62} \underline{4.68} & 3.62 & 4.35 & 3.31 & 4.03 & 4.51 & 3.08 & 3.95 & 2.60 & 3.03 & 4.40 & 2.50 & 3.34 \\
VACE-LTX  & 2B & 2.74 & 2.86 & 2.89 & 2.25 & 2.83 & 2.12 & 2.26 & 2.49 & 1.98 & 2.29 & 1.94 & 2.06 & 2.41 & 1.89 & 2.14 \\
VINO  & 13B & 3.72 & 4.22 & 4.46 & 3.42 & 4.13 & 3.56 & \color[HTML]{319B62} \underline{4.34} & \color[HTML]{319B62} \underline{4.58} & 3.39 & 4.16 & 2.63 & 2.97 & 4.24 & 2.53 & 3.28 \\
UniVideo(query)  & 13B & 3.35 & 3.90 & 4.41 & 3.01 & 3.89 & 2.98 & 3.73 & 4.18 & 2.76 & 3.63 & 2.30 & 2.50 & 3.89 & 2.20 & 2.89 \\ 
UniVideo(hidden)  & 13B & 3.33 & 4.18 & 4.38 & 3.08 & 3.97 & 3.22 & 4.12 & 4.36 & 3.09 & 3.90 & 2.31 & 2.83 & 3.94 & 2.29 & 3.03 \\ \hline
\textbf{Ours} & 8.3B & \color[HTML]{319B62} \underline{4.35} & \color[HTML]{319B62} \underline{4.53} & 4.58 & \color[HTML]{319B62} \underline{4.01} & \color[HTML]{319B62} \underline{4.49} & \color[HTML]{319B62} \underline{4.08} & 4.22 & 4.52 & \color[HTML]{319B62} \underline{3.61} & \color[HTML]{319B62} \underline{4.27} & \color[HTML]{319B62} \underline{3.53} & \color[HTML]{319B62} \underline{4.01} & \color[HTML]{319B62} \underline{4.54} & \color[HTML]{319B62} \underline{3.26} & \color[HTML]{319B62} \underline{4.03} \\
         
            \bottomrule
        \end{tabular}}
\end{table*}

\begin{table*}[t]
    \centering
    \caption{ \label{tab:edit} Main results on OpenVE-Bench. The best results are in \textbf{bold} for {\color[HTML]{F88825} specialized-} and {\color[HTML]{319B62} unified-}models, with the second best among unified models \underline{underlined}.}
    \vspace{-0.5em}
    \resizebox{1.0\linewidth}{!}{
        \begin{tabular}{lc|cccccc|c}
            \toprule
        
        \textbf{Model} & \#GenParams & Global-Style$\uparrow$ & BKG-Change$\uparrow$ & Local-Change$\uparrow$ & Local-RM$\uparrow$ & Local-Add$\uparrow$ & Subtitle-Edit$\uparrow$ & \textbf{Overall}$\uparrow$ 
        \\ \hline
        \multicolumn{9}{c}{\textit{Specialized Video Generation Models}} \\ \hline
OmniVideo  & 1.3B & 1.11 & 1.18 & 1.14 & 1.14 & 1.36 & 1.00 & 1.16 \\
InsViE  & 2B & 2.20 & 1.06 & 1.48 & 1.36 & 1.17 & 2.18 & 1.58 \\
Lucy-Edit  & 5B & 2.27 & 1.57 & \color[HTML]{F88825} \textbf{3.20} & 1.75 & \color[HTML]{F88825} \textbf{2.30} & 1.61 & 2.12 \\
ICVE   & 13B & 2.22 & 1.62 & 2.57 & \color[HTML]{F88825} \textbf{2.51} & 1.97 & 2.09 & 2.16 \\
Ditto  & 14B & \color[HTML]{F88825} \textbf{4.01} & 1.68 & 2.03 & 1.53 & 1.41 & 2.81 & 2.25 \\
OpenVE-Edit  & 5B & 3.16 & \color[HTML]{F88825} \textbf{2.36} & 2.98 & 1.85 & 2.15 & \color[HTML]{F88825} \textbf{2.91} & \color[HTML]{F88825} \textbf{2.57} \\ \hline
\multicolumn{9}{c}{\textit{Unified Video Generation Models}} \\ \hline
VACE  & 14B & 1.49 & 1.55 & 2.07 & 1.46 & 1.26 & 1.48 & 1.55 \\
VINO  & 13B & \color[HTML]{319B62} \textbf{4.32} & 2.37 & 3.65 & \color[HTML]{319B62} \underline{3.19} & 2.71 & 2.94 & \color[HTML]{319B62} \underline{3.19} \\
UniVideo(query)  & 13B & 3.47 & \color[HTML]{319B62} \textbf{2.47} & \color[HTML]{319B62} \textbf{3.86} & 2.48 & \color[HTML]{319B62} \underline{2.87} & \color[HTML]{319B62} \underline{3.13} & 3.05 \\
UniVideo(hidden)  & 13B & 3.55 & \color[HTML]{319B62} \underline{2.42} & 3.67 & 2.89 & \color[HTML]{319B62} \textbf{2.90} & 2.99 & 3.07 \\ \hline

\textbf{Ours} & 8.3B & \color[HTML]{319B62} \underline{4.16} & 2.32 & \color[HTML]{319B62} \underline{3.68} & \color[HTML]{319B62} \textbf{3.22} & 2.41* & \color[HTML]{319B62} \textbf{3.54} & \color[HTML]{319B62} \textbf{3.22} \\

            \bottomrule
        \end{tabular}}
\end{table*}

We conduct zero-shot evaluations on IntelligentVBench for both open-source task-specific models and unified models.
The primary results for  Implicit I2V, Interpolative DI2V, and TIV2V are in Table~\ref{tab:main1}, while the performance for Compositional MI2V is detailed in Table~\ref{tab:main2}.


Comparing existing academic unified models against specialized models and MiniMax-H3~\citep{minimax}, an industrial open-weight model, we identify three key observations.
\textbf{(1)} Current academic unified models exhibit significant performance imbalances across subtasks: VINO~\citep{chen2026vino} and VACE~\citep{jiang2025vace} achieve competitive results on Compositional MI2V, yet fall considerably short on Interpolative DI2V and TIV2V, respectively. Such discrepancies indicate that these models remain far from truly ``omni-capable'', lacking the robustness to handle diverse input formats while simultaneously maintaining reasoning and compositional proficiency.
\textbf{(2)} On individual subtasks, they still lag behind specialized models in reasoning-intensive scenarios, such as Implicit I2V and Interpolative DI2V. In contrast, on Compositional MI2V, they can match or even surpass specialized models. This discrepancy suggests that while current multi-task training effectively captures compositional attributes, the capacity for reasoning-informed generation remains underdeveloped.
\textbf{(3)} TIV2V, which requires the simultaneous synthesis of text, image, and video modalities, still lacks dedicated specialized models, underscoring a significant research gap in the seamless integration and coordination of increasingly diverse multimodal inputs.

Compared to existing academic unified models, OmniWeaving achieves the following advantages:
\textbf{(1) Synergistic Multi-Task Integration.} Across all four tasks, OmniWeaving consistently achieves SoTA performance among open-source academic unified models under both $\mathbf{MIN}$ and $\mathbf{AVG}$ overall metrics. 
\textbf{(2) Enhanced Multimodal Composition.} Whether processing complex multi-subject image combinations or heterogeneous tri-modal inputs (image, video, and text), OmniWeaving outperforms all academic baselines, showcasing superior compositional flexibility for video generation.
\textbf{(3) Comprehension-Guided Generation.}
For reasoning-related tasks like Implicit I2V and Interpolative DI2V, enabling the ``thinking mode'' of MLLM  yields consistently superior results compared to direct generation. This confirms that our framework successfully unifies comprehension and generation, leveraging deep reasoning to elevate visual synthesis.
\textit{Of course, compared to the industrial MiniMax-H3, which benefits from substantially larger training scale and parameter count, OmniWeaving, as an academic model, still exhibits a notable performance gap.}

\subsection{Main Results on Existing Benchmarks (T2V and V2V)}

We conduct zero-shot evaluations on VBench~\citep{huang2024vbench} for text-to-video (T2V) generation and OpenVE-Bench~\citep{he2025openve} for video-to-video (V2V) editing, comparing OmniWeaving against both unified video generation models and specialized models.

Table~\ref{tab:vbench} presents the main VBench results for text-to-video (T2V) generation, with baseline scores sourced from \cite{chen2026vino}.
We can see that OmniWeaving outperforms existing unified frameworks on VBench, achieving performance comparable to specialized models like HunyuanVideo. Given that T2V data constitutes about 10\% of our training corpus, OmniWeaving successfully preserves its foundational generation capabilities without significant degradation.

\begin{wraptable}{r}{0.5\textwidth}
    \centering
    \vspace{-1em}
    \caption{ \label{tab:vbench} Main Results on VBench. }
    
    \vspace{-0.5em}
    \resizebox{1.0\linewidth}{!}{
        \begin{tabular}{lc|cc|c}
            \toprule
               \textbf{Model} & \#GenParams & Quality$\uparrow$ & Semantic$\uparrow$
            & \textbf{Total}$\uparrow$  \\ \hline
            \multicolumn{5}{c}{\textit{Specialized Video Generation Models}} \\ \hline
            StepVideo  & 30B & 84.46 & 71.28 & 81.83 \\ 
            CogVideoX  & 5B & 83.05 & \color[HTML]{F88825} \textbf{77.33} & 81.91 \\ 
            HunyuanVideo  & 13B & 85.09 & 75.82 & 83.24 \\ 
            Wan2.1  & 14B & \color[HTML]{F88825} \textbf{85.59} & 76.11 & \color[HTML]{F88825} \textbf{83.69} \\ \hline 
            
            \multicolumn{5}{c}{\textit{Unified Video Generation Models}} \\ \hline
            VINO  & 13B & 84.00  & 78.01 & 82.80 \\ 
            UniVideo  & 13B & - & - & 82.58 \\ \hline
            \textbf{Ours} & 8.3B & \color[HTML]{319B62} \textbf{84.37} &  \color[HTML]{319B62} \textbf{78.03} &  \color[HTML]{319B62} \textbf{83.10}\\
            \bottomrule
        \end{tabular}
    }
\end{wraptable}
Table~\ref{tab:edit} presents the main V2V editing results on OpenVE-Bench. We adopt most baseline scores directly from \cite{he2025openve}, whereas the results for UniVideo and VINO are obtained through our own evaluations.
We can see that OmniWeaving yields an average score of 3.22 across the six OpenVE-Bench subtasks, surpassing specialized models and unified counterparts such as UniVideo and VINO. 
Notably, our model demonstrates well-balanced proficiency across global, local, and text-based editing. 
However, we do observe a relative performance dip in the ``local-add'' subtask. 
We find that this discrepancy arises because this subtask primarily features animated content addition with minimal emphasis on natural integration, whereas our training explicitly prioritizes realistic object addition for this editing type.

\begin{figure*}[t]
\includegraphics[width=\linewidth]{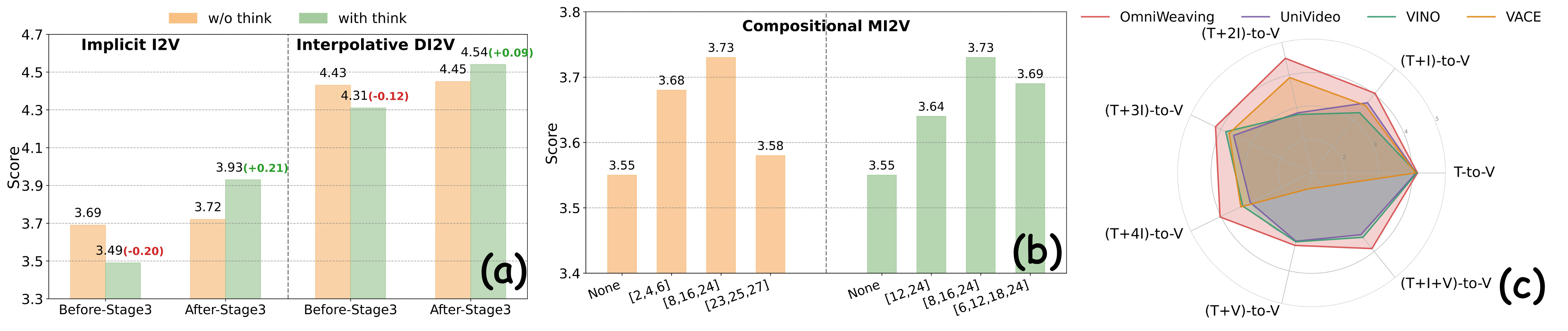}
\vspace{-1em}
\centering\caption{(a) AVG performance when enabling or disabling the thinking mode of OmniWeaving. (b) AVG performance with different DeepStacking strategies. (c) Performance visualization across different input formats for each unified video generation model.}
\label{fig:plot}
\end{figure*}

\begin{figure}[!t]
    \centering
    \includegraphics[width=\textwidth]{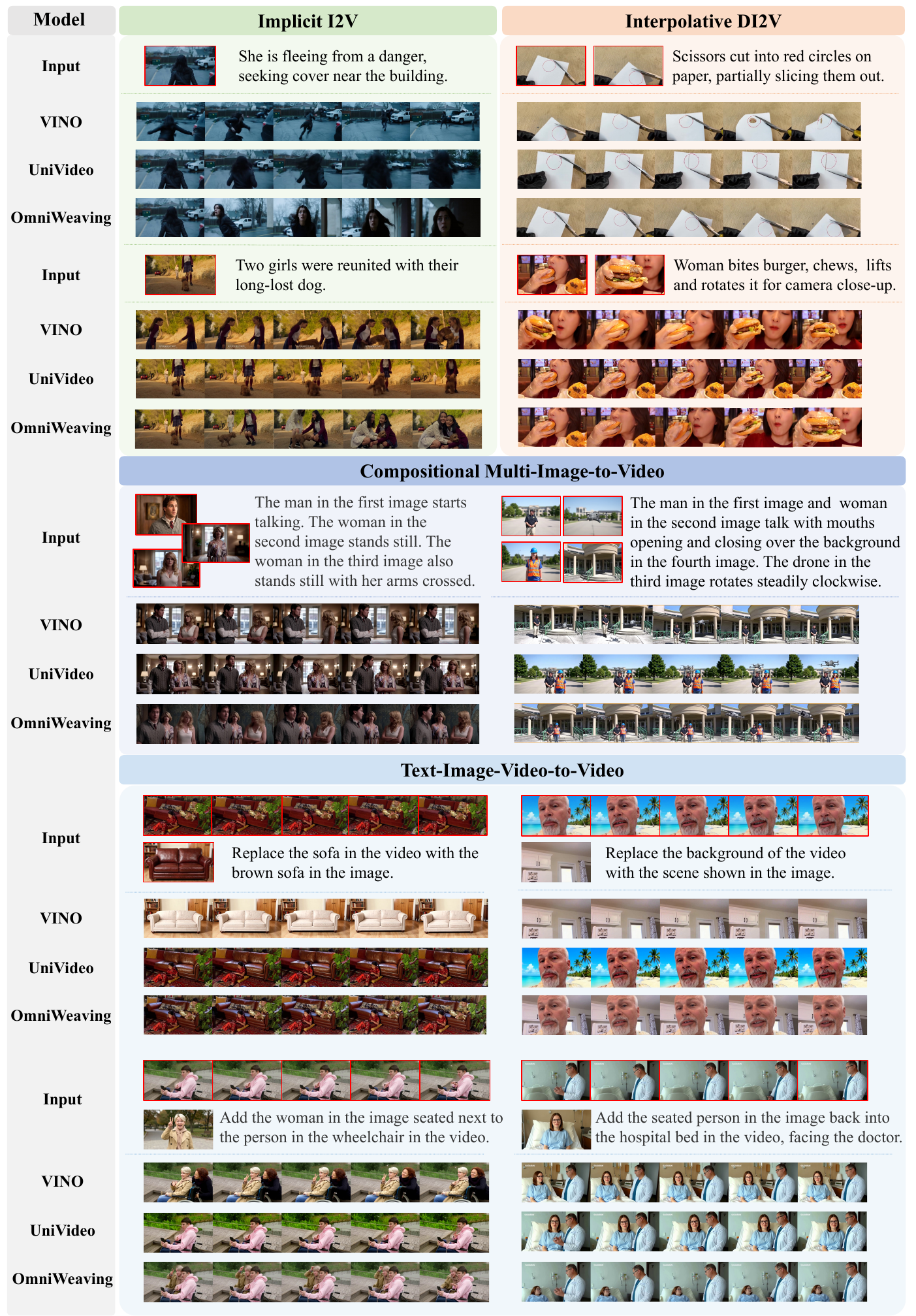} 
    \caption{Qualitative comparison of VINO, UniVideo, and OmniWeaving.}
    \label{fig:qualitative}
\end{figure}

\subsection{Qualitative Comparisons}

As illustrated in Figure~\ref{fig:qualitative}, we conduct a qualitative comparison of  VINO, UniVideo, and OmniWeaving. 
In both Implicit I2V and Interpolative DI2V tasks, the two baselines frequently produce start or end frames that are misaligned with the provided reference images. Moreover, in Compositional MI2V, both baselines struggle to successfully integrate all specified visual elements; for instance, in the first case, the videos synthesized by both VINO and UniVideo contain only two individuals, whereas in the second case, although UniVideo manages to incorporate all three subjects, it entirely disregards the background constraints. Finally, in TIV2V, the baselines often introduce unintended modifications to the unedited regions of the source video, while sometimes failing to accurately respond to the user's prompts.
In contrast, OmniWeaving effectively mitigates these issues, showing higher-quality generation across all cases. See more qualitative examples of OmniWeaving in Figures~\ref{fig:case1}, \ref{fig:case2}, \ref{fig:case3}, \ref{fig:case4}, and \ref{fig:case5} within Appendix~\ref{app:2}.

\subsection{In-Depth Analysis}
\paragraph{\textbf{Effect of Reasoning-Augmented Video Generation.}}
As shown in Figure~\ref{fig:plot}(a), we evaluate the AVG performance of OmniWeaving and its precursor prior to the final training stage on Implicit I2V and Interpolative DI2V tasks,
both with and without ``thinking mode'' activated. Notably, before training stage 3, enabling ``thinking mode'' severely degrades performance, revealing a lack of synergy between comprehension and generation. In contrast, after Reasoning-Augmented Fine-Tuning, ``thinking mode'' significantly boosts overall results, demonstrating that enhanced reasoning effectively drives higher-quality video generation.

\paragraph{\textbf{Effect of DeepStacking.}} 
To evaluate the impact of DeepStacking, we fine-tune the Stage-1 model on a multimodal-composition subset with various layer-selection strategies. Figure~\ref{fig:plot}(b) compares their average performance on Compositional MI2V. 
When fixing the selection to three layers, sampling across a broad depth (\textit{e.g.}, [8, 16, 24]) outperforms concentrations in exclusively shallow ([2, 4, 6]) or deep ([23, 25, 27]) layers, with all configurations surpassing non-DeepStacking. Furthermore, this 3-layer setup achieves superior results compared to 2-layer ([12, 24]) or 4-layer ([6, 12, 18, 24]) alternatives. These findings suggest that integrating a balanced spectrum of semantic features—spanning low-level to high-level—optimizes compositional video generation.

\begin{wraptable}{r}{0.4\textwidth}
    \centering
    \caption{ \label{tab:human}Pearson correlation with human ratings for each VLM. }
    \vspace{-0.5em}
    
    \resizebox{1.0\linewidth}{!}{
        \begin{tabular}{l|cccc}
            \toprule

            \textbf{VLMs} & $\mathcal{IF}\uparrow$ & $\mathcal{CP}\uparrow$ & $\mathcal{VQ}\uparrow$ & $\mathbf{AVG}$ \\ \hline
    Qwen3-VL-235B  & 0.66 & 0.54 & 0.59 & 0.60 \\
Seed-1.6        & 0.72 & 0.63 & 0.67 & 0.67 \\
GPT-5          & 0.75 & 0.72 & 0.71 & 0.73 \\ \hline
Gemini2.5-Pro  & \textbf{0.81} & \textbf{0.74} & \textbf{0.72} & \textbf{0.76} \\

            \bottomrule
        \end{tabular}
    }
    \vspace{-1.0em}
\end{wraptable}

\paragraph{\textbf{Capability Visualization.}} Figure~\ref{fig:plot}(c) presents a radar chart visualizing the average performance scores across diverse input formats on established test cases for each unified model. It intuitively shows that OmniWeaving consistently outperforms baselines, exhibiting superior versatility and robustness when processing a wide variety of input formats.

\paragraph{\textbf{Assessment of Evaluation Protocol.}} 
To validate the evaluation reliability for IntelligentVBench, we conducted a user study comparing human expert ratings with evaluations from several frontier VLMs, including Gemini2.5-Pro~\citep{comanici2025gemini}, GPT-5~\citep{openai2025gpt5}, Seed-1.6~\citep{guo2025seed1}, and Qwen3-VL~\citep{bai2025qwen3}. By calculating the Pearson correlation coefficient across randomly sampled representative cases from each subtask, we assessed the alignment between automated scores and human judgment. As shown in Table~\ref{tab:human}, Gemini2.5-Pro consistently achieves the highest correlation across all metrics, demonstrating its superior proficiency in accurately reflecting model performance.

\section{Conclusion and Future Work}

In this paper, we introduce OmniWeaving, an omni-level video generation model featuring powerful multimodal composition and reasoning-informed generation capabilities. 
OmniWeaving can accommodate free-form, multimodal interleaved inputs to generate compliant videos. 
It is also capable of leveraging deep visual understanding to actively guide the generative process with a unified visual comprehension and generation framework. 
Furthermore, we propose IntelligentVBench, a comprehensive benchmark specifically designed to evaluate next-level intelligent video generation, containing diverse tasks. 
Extensive experiments show that OmniWeaving achieves SoTA performance across existing open-source academic unified models and even surpasses specialized models.

Furthermore, it is imperative to clarify that OmniWeaving is not designed to entirely bridge the performance gap between academic and industrial models. We concede that a substantial disparity still persists between academic unified video generation models and industrial counterparts, including both proprietary systems like Seedance-2.0 and open-weight models like MiniMax-H3.
This gap is evident not only in overall model performance but also in the diversity of supported input modalities and format flexibility, largely because industrial models benefit from significantly greater computational resources and training data.
Closing this gap is neither our objective for OmniWeaving nor a realistic short-term expectation. 
Instead, through the exploration of OmniWeaving and its open-source release, we aim to provide the community with a viable reference point to help guide the future trajectory of unified video generation models. 
Moving forward, we will continue to explore unified video generation. 
Our future efforts will focus on supporting more complex inputs, such as interleaved multiple-image-and-video sequences, and incorporating additional modalities, such as audio input and output, to ultimately achieve fully omni-modal, audio-visually synchronized video generation.

\bibliography{iclr2025_conference}
\bibliographystyle{iclr2025_conference}
\newpage
\clearpage

\appendix

\section*{Appendix}

\section{Model Details}
\label{app:1}

Building upon the general architecture outlined in Section~\ref{sec:4.1}, we detail the specific design and implementation of our model by instantiating OmniWeaving with Qwen2.5-VL~\citep{bai2025qwen25vltechnicalreport} as the MLLM and HunyuanVideo-1.5~\citep{wu2025hunyuanvideo} as the core generative engine. Because HunyuanVideo-1.5 utilizes a multilingual Glyph-ByT5 text encoder~\citep{xue2022byt5} for precise cross-lingual text rendering and a SigLIP~\citep{zhai2023sigmoid} encoder for robust vision-text semantic alignment, we retain both modules to preserve architectural consistency with the generative backbone. Specifically, in scenarios involving multiple visual inputs, the global SigLIP feature is computed as the mean embedding across all provided images. Similarly, for video inputs, we extract a representative set of frames and average their corresponding SigLIP embeddings to yield a unified semantic representation. Notably, both the Glyph-ByT5 and SigLIP encoders remain strictly frozen throughout the entire training process.

Furthermore, a key aspect of multi-task training is that the representations provided by the MLLM should enable the MMDiT to understand exactly what task needs to be performed~\citep{wu2025qwen}. 
To this end, we set distinct MLLM system prompts for different tasks to facilitate better differentiation. 
The system prompts for each video generation task are shown in Figure~\ref{fig:sys_prompt_t2v_i2v}, Figure~\ref{fig:sys_prompt_first_last_editing}, Figure~\ref{fig:sys_prompt_composition}, and Figure~\ref{fig:sys_prompt_reasoning}.
For reasoning-augmented tasks, we append a specialized reasoning-guidance prompt to the user input (\textit{i.e.}, \textit{based on the provided image and text, please generate a detailed description that explicitly articulates the visual, temporal, and semantic characteristics of the target video}). 
This reasoning-guidance prompt explicitly directs the MLLM to perform reasoning, yielding an enhanced prompt that provides a granular description of the target video’s spatial composition and temporal dynamics.
Subsequently, the hidden states corresponding to this enhanced prompt are concatenated with those of the original user input; this combined representation is then fed into the MMDiT to effectively steer the generative process.

\section{More Experimental Results}
\label{app:2}

\begin{figure}[t]
    \centering
    \includegraphics[width=\textwidth]{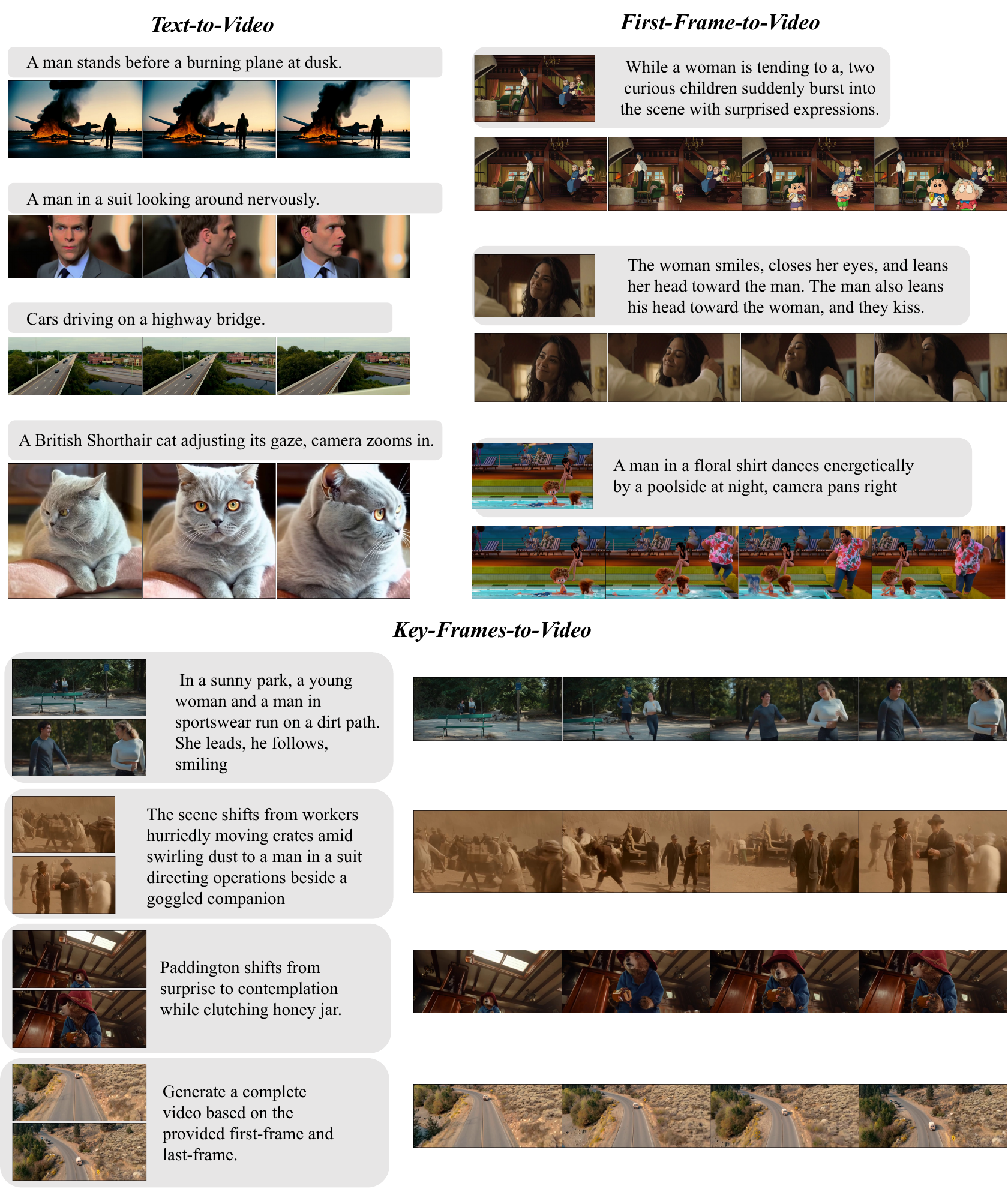} 
    \vspace{-1em}
    \caption{Qualitative results for OmniWeaving on Text-to-Video, First-Frame-to-Video, and Key-Frames-to-Video generation tasks.}
    \vspace{-2em}
    \label{fig:case1}
\end{figure}

\begin{figure}[t]
    \centering
    \includegraphics[width=\textwidth]{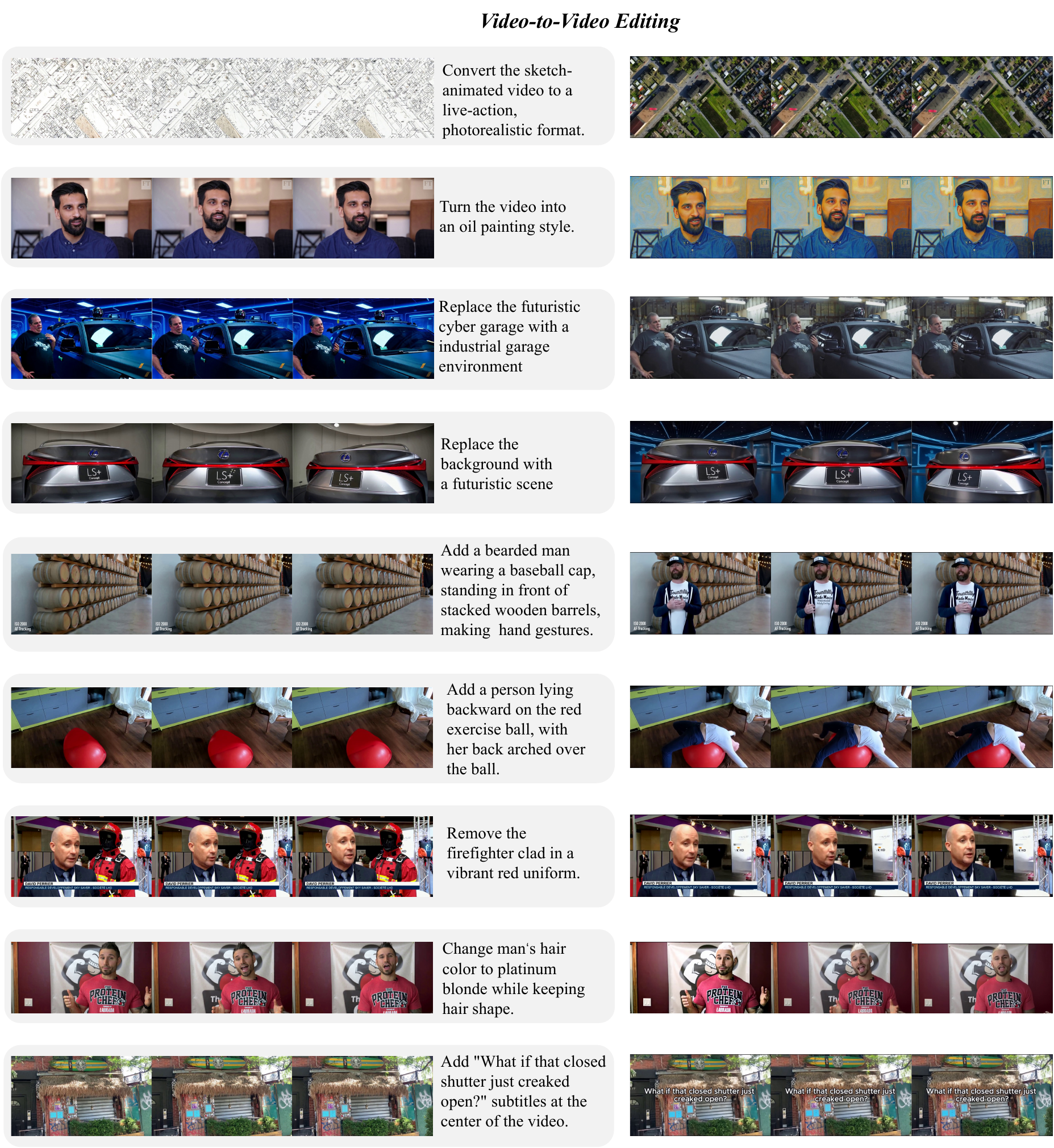} 
    \vspace{-1em}
    \caption{Qualitative results for OmniWeaving on Video-to-Video editing tasks.}
    \vspace{-2em}
    \label{fig:case2}
\end{figure}

\begin{figure}[t]
    \centering
    \includegraphics[width=\textwidth]{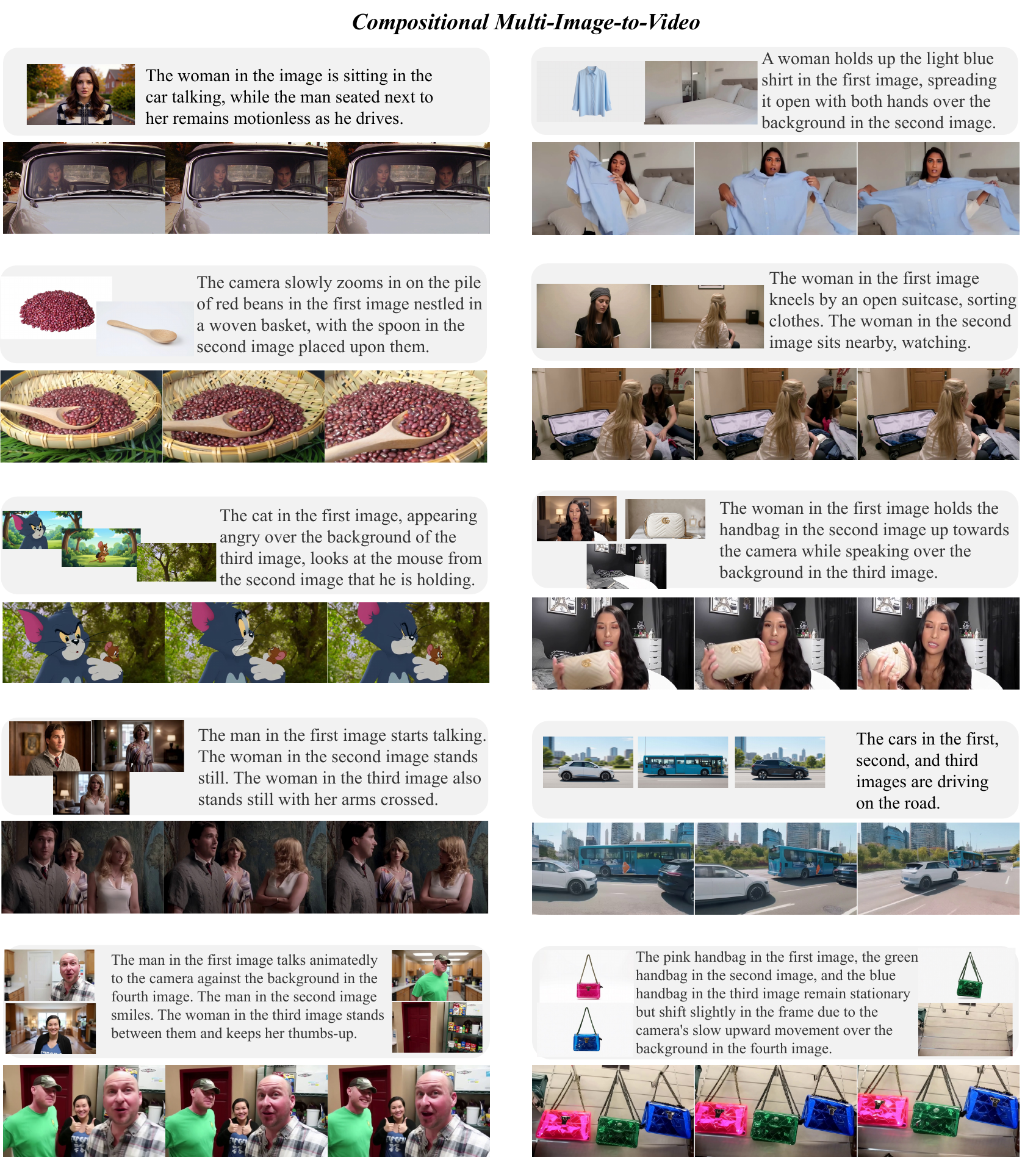} 
    \vspace{-1em}
    \caption{Qualitative results for OmniWeaving on Compositional Multi-Image-to-Video generation tasks.}
    \vspace{-2em}
    \label{fig:case3}
\end{figure}

\begin{figure}[t]
    \centering
    \includegraphics[width=\textwidth]{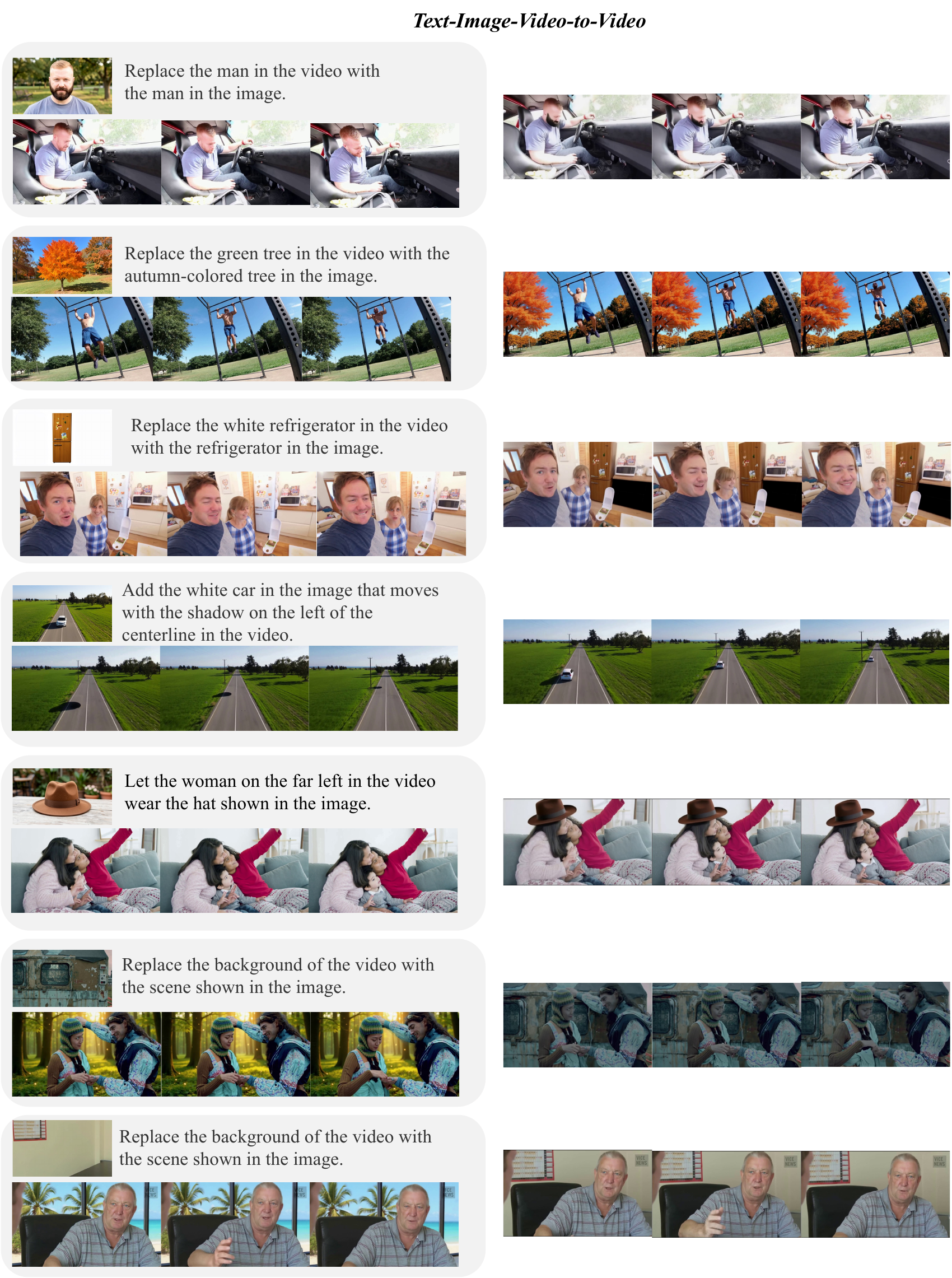} 
    \vspace{-1em}
    \caption{Qualitative results for OmniWeaving on Text-Image-Video-to-Video generation tasks.}
    \vspace{-2em}
    \label{fig:case4}
\end{figure}

\begin{figure}[t]
    \centering
    \includegraphics[width=\textwidth]{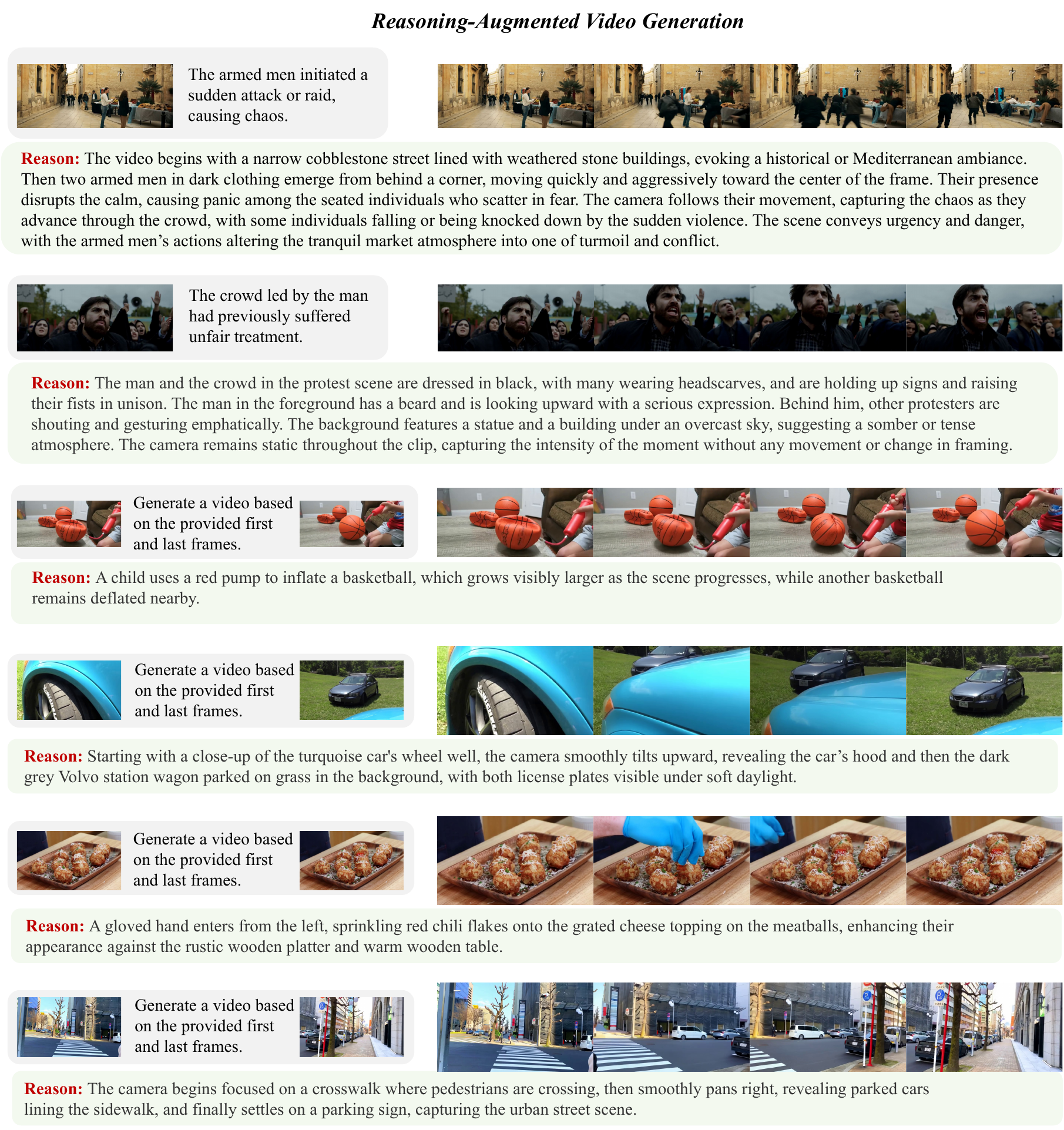} 
    \vspace{-1em}
    \caption{Qualitative results for OmniWeaving on Reasoning-Augmented video generation.}
    \vspace{-2em}
    \label{fig:case5}
\end{figure}

More qualitative results for OmniWeaving's video generation across various task scenarios are presented in Figure~\ref{fig:case1}, Figure~\ref{fig:case2}, Figure~\ref{fig:case3}, Figure~\ref{fig:case4}, and Figure~\ref{fig:case5}. As can be seen, OmniWeaving integrates diverse video generation capabilities within a single model, enabling it to effectively handle various video generation tasks.

\section{Benchmark Evaluation Prompts}
\label{app:3}

This section presents the evaluation prompt templates tailored for each task in IntelligentVBench. Overall, we utilize Gemini2.5-Pro as an automated evaluator to assess three essential metrics per task, with each metric scored on a 1–5 scale.
Specifically, the evaluation prompt template for Implicit I2V is detailed in Figure~\ref{fig:bench_i2v}.
The evaluation prompt template for Interpolative DI2V is detailed in Figure~\ref{fig:bench_di2v}.
The evaluation prompt templates for Compositional MI2V are detailed in Figure~\ref{fig:bench_mi2v} and Figure~\ref{fig:bench_mi2v_single}.
Furthermore, we design three distinct evaluation prompt templates tailored to the specific sub-tasks within the TIV2V task, as shown in Figure~\ref{fig:bench_tiv2v_add}, Figure~\ref{fig:bench_tiv2v_back}, and Figure~\ref{fig:bench_tiv2v_change}.

\begin{figure}[t]
\begin{AcademicBox}[\footnotesize System Prompt for Text-to-Video Generation Task]
\verb=<|im_start|>=system \\
You are a helpful assistant. Describe the video by detailing the following aspects: (1) The main content and theme of the video. (2) The color, shape, size, texture, quantity, text, and spatial relationships of the objects. (3) Actions, events, behaviors temporal relationships, physical movement changes of the objects. (4) Background environment, light, style and atmosphere. (5) camera angles, movements, and transitions used in the video.\verb=<|im_end|>= \\
\verb=<|im_start|>=user \\
{\color{red}\verb=<|user_text|>=}\verb=<|im_end|>= \\
\verb=<|im_start|>=assistant
\end{AcademicBox}

\begin{AcademicBox}[\footnotesize System Prompt for First-Frame-to-Video Generation Task]
\verb=<|im_start|>=system \\
You are a helpful assistant. Describe the key features of the input image (color, shape, size, texture, objects, background), then explain how the user's text instruction should alter the image to introduce motion and evolution over time. Generate a video using this image as the first frame that meets the user's requirements, ensuring the specified elements evolve or move in a way that fulfills the text description while maintaining consistency. \verb=<|im_end|>= \\
\verb=<|im_start|>=user \\
{\bfseries\color{red}\verb=<|user_img|>=\verb=<|user_text|>=}\verb=<|im_end|>= \\
\verb=<|im_start|>=assistant
\end{AcademicBox}

\vspace{-1em}
\caption{System prompts for Text-to-Video and First-Frame-to-Video generation tasks, where \textcolor{red}{\texttt{<|user\_img|>}} is the user input image and \textcolor{red}{\texttt{<|user\_text|>}} is the user input prompt.\looseness=-1}
\label{fig:sys_prompt_t2v_i2v}
\end{figure}
\begin{figure}[t]

\begin{AcademicBox}[\footnotesize System Prompt for Key-Frames-to-Video Generation Task]
\verb=<|im_start|>=system \\
You are a helpful assistant. Given a text instruction and multiple images (greater than 1) as key-frames of the video, you need to analyze the visual trajectory required to transition from the start to the end across these key-frames. Determine how the elements in each key-frame must evolve, move, or transform to align with the subsequent key-frame based on the text instruction. Generate a video that seamlessly connects these key-frames, ensuring the motion and evolution between them fulfill the text description while maintaining temporal consistency. \verb=<|im_end|>= \\
\verb=<|im_start|>=user \\
{\bfseries\color{red}\verb=<|key_frame_1|>=...\verb=<|key_frame_n|>=\verb=<|user_text|>=}\verb=<|im_end|>= \\
\verb=<|im_start|>=assistant
\end{AcademicBox}

\begin{AcademicBox}[\footnotesize System Prompt for Video-to-Video Editing Task]
\verb=<|im_start|>=system \\
You are a helpful assistant. Given a text instruction and an input video, you need to analyze the visual content and temporal dynamics of the input video, and then explain how the user's text instruction should modify the video's visual style, objects, or scene composition. Generate an edited video that meets the user's requirements, ensuring the specified modifications are applied consistently across frames while preserving the original motion flow and coherence.\verb=<|im_end|>= \\
\verb=<|im_start|>=user \\
{\bfseries\color{red}\verb=<|user_video|>=\verb=<|user_text|>=}\verb=<|im_end|>= \\
\verb=<|im_start|>=assistant
\end{AcademicBox}

\vspace{-1em}
\caption{System prompts for Key-Frames-to-Video generation and Video-to-Video editing tasks, where \textcolor{red}{\texttt{<|user\_text|>}} is the user input prompt, \textcolor{red}{\texttt{<|key\_frame\_i|>}} is the user input image, and \textcolor{red}{\texttt{<|user\_video|>}} is the user input video.\looseness=-1}
\label{fig:sys_prompt_first_last_editing}
\vspace{-1em}

\end{figure}
\begin{figure}[t]

\begin{AcademicBox}[\footnotesize System Prompt for Interleaved Text-and-Multi-Image-to-Video Generation Task]
\verb=<|im_start|>=system \\
You are a helpful assistant. Given text instructions and one or more input images, you need to explain how to extract and combine key information from the input images to construct a new image as the video's first frame, and then how the user's text instruction should alter the image to introduce motion and evolution over time. Generate a video that meets the user's requirements, ensuring the specified elements evolve or move in a way that fulfills the text description while maintaining consistency.\verb=<|im_end|>= \\
\verb=<|im_start|>=user \\
{\bfseries\color{red}\verb=<|user_img_1|>=\verb=<|user_text_1|>=...\verb=<|user_img_n|>=\verb=<|user_text_n|>=}\verb=<|im_end|>= \\
\verb=<|im_start|>=assistant
\end{AcademicBox}

\begin{AcademicBox}[\footnotesize System Prompt for Text-Image-Video-to-Video Generation Task]
\verb=<|im_start|>=system \\
You are a helpful assistant. Given a text instruction, one or more reference images, and an input video, you need to analyze the visual content and temporal dynamics of the input video, alongside the scene or subject characteristics of the reference images. Explain how the user's text instruction directs the application of the reference images' visual attributes onto the input video. Generate an edited video that meets the user's requirements, ensuring the specified modifications are applied consistently across frames while preserving the original motion flow and coherence. \verb=<|im_end|>= \\
\verb=<|im_start|>=user \\
{\bfseries\color{red}\verb=<|user_video|>=\verb=<|user_img_1|>=...\verb=<|user_img_n|>=\verb=<|user_text|>=}\verb=<|im_end|>= \\
\verb=<|im_start|>=assistant
\end{AcademicBox}

\vspace{-1em}
\caption{System prompts for Interleaved Text-and-Multi-Image-to-Video and Text-Image-Video-to-Video generation tasks, where \textcolor{red}{\texttt{<|user\_img\_i|>}} is the user input images, and \textcolor{red}{\texttt{<|user\_video|>}} is the user input video.\looseness=-1}
\label{fig:sys_prompt_composition}
\end{figure}

\begin{figure}[t]

\begin{AcademicBox}[\footnotesize System Prompt for Reasoning-augmented Task (Example 1)]
\verb=<|im_start|>=system \\ 
You are a helpful assistant. Describe the key features of the input image (color, shape, size, texture, objects, background), then explain how the user's text instruction should alter the image to introduce motion and evolution over time. Note that the text instruction may implicitly express the intent of video generation. Generate a video using this image as the first frame that meets the user's requirements, ensuring the specified elements evolve or move in a way that fulfills the text description while maintaining consistency. \verb=<|im_end|>= \\
\verb=<|im_start|>=user \\
{\bfseries\color{red}\verb=<|user_img|>=\verb=<|user_text|>=}\\
\textcolor{red}{Based on the provided image and text, please generate a detailed description that explicitly articulates the visual, temporal, and semantic characteristics of the target video.}\verb=<|im_end|>= \\
\verb=<|im_start|>=assistant
\end{AcademicBox}

\begin{AcademicBox}[\footnotesize System Prompt for Reasoning-augmented Task (Example 2)]
\verb=<|im_start|>=system \\ 
You are a helpful assistant. Given a text instruction and multiple images (greater than 1) as key-frames of the video, you need to analyze the visual trajectory required to transition from the start to the end across these key-frames. Determine how the elements in each key-frame must evolve, move, or transform to align with the subsequent key-frame based on the text instruction. Note that the text instruction may implicitly express the intent of video generation. Generate a video that seamlessly connects these key-frames, ensuring the motion and evolution between them fulfill the text description while maintaining temporal consistency. \verb=<|im_end|>= \\
\verb=<|im_start|>=user \\
{\bfseries\color{red}\verb=<|key_frame_1|>...<|key_frame_n|><|user_text|>=}\\
\textcolor{red}{Based on the provided image and text, please generate a detailed description that explicitly articulates the visual, temporal, and semantic characteristics of the target video.}\verb=<|im_end|>= \\
\verb=<|im_start|>=assistant
\end{AcademicBox}

\vspace{-1em}
\caption{Two system prompt examples for reasoning-augmented video generation tasks.\looseness=-1}
\label{fig:sys_prompt_reasoning}
\vspace{-1em}
\end{figure}

\begin{figure}[t]

\begin{AcademicBox}[ Evaluation Prompt Template for Implicit I2V Task]
You are an expert data rater specializing in evaluating Image-to-Video (I2V) generation. You will be provided with a reference image (serving as the first frame), a natural language instruction, and the generated video. 
Note that the instruction may describe abstract intent, introduce new characters or objects not present in the input image, or may lack an explicit association with the input image. Your task is to evaluate the generated video on a 5-point scale across three dimensions: \\
\textbf{1. The first score: Condition Preserving} \\
\textbf{Objective:} Evaluate whether the first frame of the generated video perfectly anchors to the input image. Maintaining style consistency and coherence in subsequent frames relative to the first frame is considered a secondary, advanced requirement for achieving the highest scores. \\
\textbf{- 5:} Perfect Consistency. The first frame is an identical reconstruction of the input image. Additionally, as an advanced requirement, subsequent frames maintain impeccable style consistency and visual coherence with the first frame without abrupt mutations.
\textbf{- 4:} High Fidelity. The first frame is highly consistent with the input image, with only negligible differences (e.g., micro-scaling or slight cropping). Minor stylistic drifting or small visual jumps in subsequent frames are acceptable, provided the first frame is nearly perfect.
\textbf{- 3:} Moderate Consistency. The first frame is recognizable as the input image but shows visible shifts in color, sharpness, or structural details. The score drops here primarily because the starting point is flawed, making subsequent frame behavior less relevant.
\textbf{- 2:} Low Fidelity. The first frame deviates significantly from the input (e.g., distorted composition, major style shifts, or missing key elements). The sequence is heavily penalized due to the initial frame mismatch, regardless of what happens next.
\textbf{- 1:} Total Dissociation. The video completely fails to use the input image as the first frame. There is zero visual connection to the reference from the very beginning. \\
\textbf{2. The second score: Instruction Following} \\
\textbf{Objective:} Assess how well the video executes the semantic intent of the prompt, including the "silky" integration of new elements and the logical evolution of the scene. \\
\textbf{- 5:} Perfect Alignment. Every detail of the instruction is perfectly rendered. If new characters or objects are introduced, they appear through logical, seamless transitions rather than abrupt popping. The semantic progression is intuitive and creative.
\textbf{- 4:} Good Alignment. The core semantics of the prompt are successfully captured. New elements are integrated well, though there might be minor imperfections in the pacing of the motion or slight omissions of secondary descriptors.
\textbf{- 3:} Partial Alignment. The model captures the main idea but ignores specific nuances. Transitions involving new elements may feel slightly awkward or forced, and the video evolution only partially matches the intended narrative.
\textbf{- 2:} Weak Alignment. The video mostly ignores the prompt or only reflects isolated keywords. The movement or introduction of elements feels chaotic, illogical, or contradicts the instruction's intent.
\textbf{- 1:} No Alignment. The video content has no relevance to the natural language instruction provided. \\
\textbf{3. The third score: Overall Visual Quality} \\
\textbf{Objective:} Evaluate the general aesthetic quality, temporal consistency (flicker-free), motion smoothness, and physical naturalness (absence of severe artifacts or morphing) of the generated video, independent of the input image or prompt. \\
\textbf{- 5:} Excellent. Exceptional visual aesthetics and perfect temporal consistency. Motions are fluid and natural (no jitter or dropped frames), physical transformations obey real-world physics, and there is zero flickering or artifacts.
\textbf{- 4:} Good. High overall quality. Smooth and coherent, with only minor, easily forgivable artifacts or very slight flickering occurring in complex motion areas or fine edges. Overall viewing experience remains pleasant.
\textbf{- 3:} Fair. Acceptable but noticeably flawed. Visible visual anomalies are present, such as noticeable flickering, mild stuttering/jittering, moderate morphological distortions during movement, or obvious AI artifacts.
\textbf{- 2:} Poor. Significantly degraded quality. Severe visual and temporal defects dominate the video, including high-frequency flickering, severe motion rigidity, grotesque "melting" or distortion of objects during transitions, and heavy noise.
\textbf{- 1:} Unacceptable. Completely collapsed visual integrity. The video lacks any temporal coherence. It is heavily corrupted with extreme artifacts, tearing, or noise, making it unwatchable and devoid of aesthetic value. \\

\textbf{Example Response Format:}
You are required to return a dictionary structured as follows: \{"Condition Preserving": [A number from 1 to 5], "Instruction Following": [A number from 1 to 5], "Overall Visual Quality": [A number from 1 to 5]\}. \\

{\bfseries\color{red}\verb=<|user_input|>=}\\

\end{AcademicBox}
\vspace{-1em}
\caption{Evaluation prompt template for Implicit I2V task.\looseness=-1}
\label{fig:bench_i2v}
\end{figure}

\begin{figure}[t]

\begin{AcademicBox}[ Evaluation Prompt Template for Interpolative DI2V Task]

You are an expert data rater specializing in evaluating video generation conditioned on first and last frames. You will be provided with two reference images, the target first frame and the target last frame, along with a text instruction, and the generated video. The instruction requires the video to transition logically from the first frame to the last frame based on the provided text. Your task is to evaluate the generated video on a 5-point scale from three perspectives: \\
\textbf{1. The first score: Condition Preserving} \\
\textbf{Objective:} Evaluate how accurately the generated video's first and last frames reconstruct the two provided input images (including color, lighting, composition, and fine details). \\
\textbf{- 5:} Perfect Match. The first and last frames of the generated video perfectly replicate the provided images. There is no perceptible cropping, or color shift.
\textbf{- 4:} High Fidelity. The first and last frames are highly consistent with the provided images. The main subjects are identical, with only microscopic, easily overlooked flaws (e.g., slight lighting variance or minuscule distortion at the extreme edges).
\textbf{- 3:} Moderate Fidelity. The first and last frames are generally recognizable as the input images but contain visible discrepancies. Noticeable issues such as mild color distortion, blurred background details, or slight morphological changes are present, though the core identity remains intact.
\textbf{- 2:} Low Fidelity. The first or the frame shows severe deviation from the provided image. There is significant distortion, missing key elements, or massive color shifts. The original image traits are barely identifiable.
\textbf{- 1:} Complete Failure. The first or last frame is completely irrelevant to the provided image, or the video fails to anchor to the first frame or the last frame entirely. \\
\textbf{2. The second score: Instruction Following} \\
\textbf{Objective:} Assess how well the generated video follows the natural language instruction to logically and semantically bridge the first and last frames. \\
\textbf{- 5:} Perfect Alignment. The transition logic from the first frame to the last frame in the generated video is flawless and highly intuitive. The video executes the text prompt perfectly. The bridge between the first and the last frames is seamless with no logical gaps.
\textbf{- 4:} Good Alignment: The transition from the first frame to the last frame is logical and cohesive. The core semantics of the prompt are successfully rendered, but there may be minor imperfections in secondary motion details, pacing, or amplitude.
\textbf{- 3:} Partial Alignment. The transition from the first frame to the last frame is somewhat disjointed. The model clearly attempts to follow the prompt to connect the keyframes, but exhibits noticeable logical leaps, introduces unwarranted hallucinatory actions, or ignores a portion of the prompt.
\textbf{- 2:} Weak Alignment. The transition from the first frame to the last frame is abrupt or chaotic. Most of the text instructions are ignored. The connection between keyframes feels like an unnatural cross-fade (morphing without physical logic), or the semantic evolution strongly violates common sense.
\textbf{- 1:} No Alignment: The video completely ignores the prompt. The intermediate generation deviates entirely from the text description, or the semantic logic is fundamentally broken and contradictory to the instruction.\\
\textbf{3. The third score: Overall Visual Quality} \\
\textbf{Objective:} Evaluate the general aesthetic quality, temporal consistency (flicker-free), motion smoothness, and physical naturalness (absence of severe artifacts or morphing) of the generated video, independent of the input images or prompt. \\
\textbf{- 5:} Excellent. Exceptional visual aesthetics and perfect temporal consistency. Motions are fluid and natural (no jitter or dropped frames), physical transformations obey real-world physics, and there is zero flickering or artifacts.
\textbf{- 4:} Good. High overall quality. Smooth and coherent, with only minor, easily forgivable artifacts or very slight flickering occurring in complex motion areas or fine edges. Overall viewing experience remains pleasant.
\textbf{- 3:} Fair. Acceptable but noticeably flawed. Visible visual anomalies are present, such as noticeable flickering, mild stuttering/jittering, moderate morphological distortions during movement, or obvious AI artifacts.
\textbf{- 2:} Poor. Significantly degraded quality. Severe visual and temporal defects dominate the video, including high-frequency flickering, severe motion rigidity, grotesque "melting" or distortion of objects during transitions, and heavy noise.
\textbf{- 1:} Unacceptable. Completely collapsed visual integrity. The video lacks any temporal coherence. It is heavily corrupted with extreme artifacts, tearing, or noise, making it unwatchable and devoid of aesthetic value. \\

\textbf{Example Response Format:}
You are required to return a dictionary structured as follows: \{"Condition Preserving": [A number from 1 to 5], "Instruction Following": [A number from 1 to 5], "Overall Visual Quality": [A number from 1 to 5]\}. \\

{\bfseries\color{red}\verb=<|user_input|>=}\\

\end{AcademicBox}

\vspace{-1em}
\caption{Evaluation prompt template for Interpolative DI2V task.\looseness=-1}
\label{fig:bench_di2v}
\end{figure}

\begin{figure}[t]

\begin{AcademicBox}[ Evaluation Prompt Template for Compositional MI2V Task with One Input Image]
You are an expert data rater specializing in evaluating subject-driven video generation. You will be given a reference image containing a specific subject, a prompt, and the generated video. The prompt requires the generated video to accurately feature the exact subject from the reference image while performing actions or interacting with the environment based on the provided text. Your task is to evaluate the generated video on a 5-point scale from three perspectives: \\
\textbf{1. The first score: Condition Preserving} \\
\textbf{Objective:} Evaluate how accurately the generated video retains the identity of the subject from the reference image throughout the entire video duration. \\
\textbf{- 5:} Perfect Preservation. The subject's identity are flawlessly maintained and highly stable. Regardless of camera angle or complex motion, the subject exactly matches the reference image with zero feature loss or morphological distortion.
\textbf{- 4:} High Preservation. The subject is highly consistent. The identity is clearly recognizable, with easily overlooked losses of detail or slight warping occurring during complex trajectories.
\textbf{- 3:} Moderate Preservation. The subject is generally recognizable but exhibits noticeable flaws. Visible loss of fine detail, color shifts, or moderate morphological distortions (e.g., slight "melting" or facial warping during movement) are present.
\textbf{- 2:} Low Preservation. The subject suffers from severe identity drift. While there might be a vague resemblance initially, the subject undergoes massive distortion, grotesque AI morphing, or noticeably transforms into a different entity as the video progresses.
\textbf{- 1:} Complete Failure. No relation to the reference. The subject in the generated video is completely irrelevant to the reference image, or the requested subject is entirely absent from the scene. \\
\textbf{2. The second score: Instruction Following} \\
\textbf{Objective:} Assess how accurately the video executes the text prompt, such as the subject's actions, the surrounding environment, object interactions, and camera movements. \\
\textbf{- 5:} Perfect Alignment. The video is a flawless semantic match to the text. It accurately executes the core actions and perfectly captures all described background settings, lighting constraints, secondary objects, and specific cinematography (e.g., panning, zooming), provided these elements are described in the prompt.
\textbf{- 4:} Good Alignment. The core intent is successfully rendered. The subject performs the main instructed actions in the correct setting, but the video misses minor secondary details (e.g., a small prop).
\textbf{- 3:} Partial Alignment. The execution is somewhat disjointed. The video captures the general concept but exhibits flawed action logic, fails to complete the motion, or entirely misses a significant environmental or action constraint mentioned in the prompt.
\textbf{- 2:} Weak Alignment. The video severely deviates from the instruction. It only captures isolated keywords (e.g., the subject is present but doing the wrong thing), and the scenario or behavior strongly contradicts the prompt's description.
\textbf{- 1:} No Alignment. Complete ignorance of the prompt. Aside from potentially including the subject, the events, actions, or environments in the video have absolutely no relation to the text prompt, representing a total hallucination. \\
\textbf{3. The third score: Overall Visual Quality} \\
\textbf{Objective:} Evaluate the general aesthetic quality, temporal consistency (flicker-free), motion smoothness, and physical naturalness of the generated video. \\
\textbf{- 5:} Excellent. Exceptional visual aesthetics and perfect temporal consistency. Motions are fluid and natural (no jitter or dropped frames), physical transformations obey real-world physics, and there is zero flickering or artifacts.
\textbf{- 4:} Good. High overall quality. Smooth and coherent, with only minor, easily forgivable artifacts or very slight flickering occurring in complex motion areas or fine edges. Overall viewing experience remains pleasant.
\textbf{- 3:} Fair. Acceptable but noticeably flawed. Visible visual anomalies are present, such as noticeable flickering, mild stuttering/jittering, moderate morphological distortions during movement, or obvious AI artifacts.
\textbf{- 2:} Poor. Significantly degraded quality. Severe visual and temporal defects dominate the video, including high-frequency flickering, severe motion rigidity, grotesque "melting" or distortion of objects during transitions, and heavy noise.
\textbf{- 1:} Unacceptable. Completely collapsed visual integrity. The video lacks any temporal coherence. It is heavily corrupted with extreme artifacts, tearing, or noise, making it unwatchable and devoid of aesthetic value.
 \\
 
\textbf{Example Response Format:}
You are required to return a dictionary structured as follows: \{"Condition Preserving": [A number from 1 to 5], "Instruction Following": [A number from 1 to 5], "Overall Visual Quality": [A number from 1 to 5]\}. \\

{\bfseries\color{red}\verb=<|user_input|>=}\\

\end{AcademicBox}
\vspace{-1em}
\caption{Evaluation prompt template for Compositional  MI2V task with one input image.\looseness=-1}
\label{fig:bench_mi2v_single}
\end{figure}

\begin{figure}[t]

\begin{AcademicBox}[ Evaluation Prompt Template for Compositional  MI2V Task with Multiple Input Images]
You are an expert data rater specializing in evaluating multi-subject-and-background-conditioned video generation. You will be given multiple reference images containing specific subjects, a reference background image, a prompt, and the generated video. The prompt requires the generated video to seamlessly integrate all exact subjects from the references into the specified background, while performing dynamic actions and interactions based on the provided text. Your task is to evaluate the video on a 5-point scale from three perspectives: \\
\textbf{1. The first score: Condition Preserving} \\
\textbf{Objective:} Evaluate how accurately the generated video retains the identity of all subjects and the background from the reference images, maintaining their stability throughout the video without identity bleeding or missing subjects. \\
\textbf{- 5:} Perfect Preservation. All subjects' identity are flawlessly maintained and highly stable. Regardless of camera angle or complex motion, each subject exactly matches the reference image with zero feature loss or morphological distortion.
\textbf{- 4:} High Preservation. All subjects and background are highly consistent with those in the given images. There is only microscopic detail loss in subjects or minuscule blurring in extreme background areas during highly complex occlusions or sweeping camera movements.
\textbf{- 3:} Moderate Preservation. Generally recognizable but flawed. No subjects are missing, and the background matches the reference broadly. However, visible issues exist: mild identity bleeding between subjects, or the background undergoes moderate morphological changes over time (e.g., specific background elements shifting, mutating, or disappearing).
\textbf{- 2:} Low Preservation. The video omits a required subject, exhibits severe identity fusion, or the background significantly deviates from the provided reference (e.g., the structural layout collapses or morphs into a different environment entirely).
\textbf{- 1:} Complete Failure. Irrelevant or severe omissions. The majority of subjects are missing, and the background bears no resemblance to the provided reference, with a total loss of input. \\
\textbf{2. The second score: Instruction Following} \\
\textbf{Objective:} Assess how accurately the video executes the text prompt, focusing on the specific actions of each subject and the logical interactions between them within the background. \\
\textbf{- 5:} Perfect Alignment. The video is a flawless semantic match to the text. It accurately executes the core actions within the specific background. Also, the multi-subject interactions described in the prompt (e.g., conversing, holding hands, fighting) are rendered perfectly and logically.
\textbf{- 4:} Good Alignment. The core intent is successfully rendered. Main interactions and actions are executed correctly within the background, but the video misses minor secondary behavioral details or subtle narrative elements.
\textbf{- 3:} Partial Alignment. Execution is somewhat disjointed. The video captures the concept of the subjects being together, but the interactions are flawed or stiff (e.g., prompting "hugging" but generating them just "standing next to each other"), or a major action constraint is missed.
\textbf{- 2:} Weak Alignment. The video severely deviates from the instruction. The model captures the presence of the subjects but completely fails to demonstrate the requested interaction or actions, with subjects acting independently and contrary to the prompt.
\textbf{- 1:} No Alignment. Complete ignorance of the prompt. Aside from containing the subjects, the events, actions, or environments have absolutely no relation to the text prompt, representing a total AI hallucination. \\
\textbf{3. The third score: Overall Visual Quality} \\
\textbf{Objective:} Evaluate the general aesthetic quality, temporal consistency (flicker-free), motion smoothness, and physical naturalness of the generated video. \\
\textbf{- 5:} Excellent. Exceptional visual aesthetics and perfect temporal consistency. Motions are fluid and natural (no jitter or dropped frames), physical transformations obey real-world physics, and there is zero flickering or artifacts.
\textbf{- 4:} Good. High overall quality. Smooth and coherent, with only minor, easily forgivable artifacts or very slight flickering occurring in complex motion areas or fine edges. Overall viewing experience remains pleasant.
\textbf{- 3:} Fair. Acceptable but noticeably flawed. Visible visual anomalies are present, such as noticeable flickering, mild stuttering/jittering, moderate morphological distortions during movement, or obvious AI artifacts.
\textbf{- 2:} Poor. Significantly degraded quality. Severe visual and temporal defects dominate the video, including high-frequency flickering, severe motion rigidity, grotesque "melting" or distortion of objects during transitions, and heavy noise.
\textbf{- 1:} Unacceptable. Completely collapsed visual integrity. The video lacks any temporal coherence. It is heavily corrupted with extreme artifacts, tearing, or noise, making it unwatchable and devoid of aesthetic value. \\
 
\textbf{Example Response Format:}
You are required to return a dictionary structured as follows: \{"Condition Preserving": [A number from 1 to 5], "Instruction Following": [A number from 1 to 5], "Overall Visual Quality": [A number from 1 to 5]\}. \\

{\bfseries\color{red}\verb=<|user_input|>=}\\

\end{AcademicBox}
\vspace{-1em}
\caption{Evaluation prompt template for Compositional MI2V task with multiple input images.\looseness=-1}
\label{fig:bench_mi2v}
\end{figure}

\begin{figure}[t]

\begin{AcademicBox}[Evaluation Prompt Template for TIV2V Task (local object addition)]
You are a data rater specializing in grading video object addition editing. You will be given two videos (before and after editing), an reference image and the corresponding editing instructions. The instructions describe how to add a specific subject from the reference image into a designated location within the video. Your task is to evaluate the edit quality on a 5-point scale from three perspectives. \\
\textbf{The first score: Instruction Following} \\
\textbf{Objective:} Evaluates whether the edit correctly follows the instruction to add the subject from the reference image into the designated location, and whether the newly added object naturally integrates into the original scene rather than looking like a rigid insertion. \\
\textbf{- 1:} Fail. No edit performed, the video is completely corrupted, or the edit is fundamentally wrong.
\textbf{- 2:} Poor. Wrong object/class added, the target is only partially added, an unrelated object is added alongside it, or it is placed in a completely incorrect location.
\textbf{- 3:} Fair. Correct object added to the general location, but with significant attribute errors (e.g., identity mismatch with the reference image). Integration check: The object feels rigidly or stiffly "pasted" into the environment and does not logically or naturally fit the context of the original scene.
\textbf{- 4:} Good. The correct object is added with main attributes and location correct; only minor details slightly mismatch the reference image. Integration check: The object integrates almost naturally into the original scene context, with only minor signs of artificial placement.
\textbf{- 5:} Perfect. All and only the requested objects are added exactly as instructed. The object perfectly matches the reference image, is placed in the exact correct position, and is integrated so naturally that it looks as though it was always meant to be part of the scene. \\
\textbf{The second score: Condition Preserving} \\
\textbf{Objective:} Evaluates two critical preservation aspects: (A) How well the newly added object retains the identity and intricate details of the subject in the reference image, and (B) How perfectly the unedited regions, non-target objects, and the original temporal dynamics from the original video are preserved. \\
\textbf{- 1:} Fail. The new object bears no resemblance to the reference image, OR the unedited regions/temporal dynamics are completely destroyed or frozen.
\textbf{- 2:} Poor. The added object lacks key intricate details from the reference image, OR there are obvious, distracting alterations to the unedited regions in the video or their original temporal flow (e.g., previously moving unedited regions suddenly freeze).
\textbf{- 3:} Fair. The added object retains its basic identity but loses some finer details from the reference. The unedited regions in the video are mostly preserved, but some spatial details or original temporal variations (e.g., altered camera shifts or smoothed-out motion in the unedited regions) are noticeably affected.
\textbf{- 4:} Good. The added object matches the reference image closely with only minor detail discrepancies. The unedited regions in the video and their temporal dynamics are almost maintained, with only minor deviations visible.
\textbf{- 5:} Perfect. Flawless preservation. The new object exactly matches the reference subject's details, AND NO changes occurred in the unedited areas, e.g., the original background and its temporal changes/motion are perfectly intact. \\
\textbf{The third score: Overall Visual Quality} \\ 
\textbf{Objective:} Evaluates the physical realism, temporal stability, and visual harmony of the new object within the video. CRITICAL EXCEPTION: Do NOT deduct points for visual quality issues, noise, or artifacts that were already present in the original, unedited video. \\
\textbf{- 1:} Fail. Severe new artifacts or physical errors (e.g., the added object floats, has completely wrong perspective/lighting). The added object causes severe flickering/jittering or blocks key original elements, breaking visual coherence.
\textbf{- 2:} Poor. Obvious paste marks, severe style/resolution mismatches, or poor handling of physics (e.g., terrible occlusion, lack of contact with surfaces). The added region temporally jitters or remains static against a moving environment.
\textbf{- 3:} Fair. The general style, lighting, and physics of the added object are fundamentally consistent with the scene, but noticeable flaws remain (e.g., slight temporal disharmony, mismatched shadows, awkward motion handling, or minor edge artifacts).
\textbf{- 4:} Good. Style, lighting, shadows, and reflections of the new object are believable and move correctly with the scene. Minor physical mismatches or slight temporal instability (e.g., faint flicker) are visible.
\textbf{- 5:} Perfect. Seamlessly integrated and physically flawless. The added object exhibits precise highlights, shadows, and motion effects that perfectly match the scene's physics. It is temporally stable and visually indistinguishable from a real object interacting within that specific environment. \\

\textbf{Example Response Format:}
You are required to return a dictionary structured as follows: \{"Condition Preserving": [A number from 1 to 5], "Instruction Following": [A number from 1 to 5], "Overall Visual Quality": [A number from 1 to 5]\}. \\

{\bfseries\color{red}\verb=<|user_input|>=}\\

\end{AcademicBox}
\vspace{-1em}
\caption{Evaluation prompt template for TIV2V task (local object addition).\looseness=-1}
\label{fig:bench_tiv2v_add}
\end{figure}

\begin{figure}[t]

\begin{AcademicBox}[ Evaluation Prompt Template for TIV2V task (background change)]
You are a data rater specializing in grading video background editing. You will be given two videos (before and after editing), an reference image and the corresponding editing instructions. The instructions requires to replace the video background with the scene depicted in the reference image. Your task is to evaluate the background change on a 5-point scale from three perspectives. \\
\textbf{The first score: Instruction Following} \\
\textbf{Objective:} Evaluates whether the background is correctly replaced and matches the target scene provided in the reference image. \\
\textbf{- 1:} Fail. No background change at all, or the new background is entirely unrelated to the reference image.
\textbf{- 2:} Poor. Background is only partially replaced, or exhibits major deviations in style, content, or layout compared to the reference image.
\textbf{- 3:} Fair. The main background is replaced, but some key elements from the reference image are missing, extraneous, or placed incorrectly.
\textbf{- 4:} Good. Requested background is fully present and largely consistent with the reference image in content and style, with only minor discrepancies.
\textbf{- 5:} Perfect. Perfect compliance; the new background perfectly matches the scene shown in the reference image (content, style, and layout). \\
\textbf{The second score: Condition Preserving} \\ 
\textbf{Objective:} Evaluates whether the non-background elements (foreground subjects, objects) and their original motion are kept completely intact without distortion. \\
\textbf{- 1:} Fail. Massive distortion, loss of foreground objects, or original motion is completely ruined/frozen.
\textbf{- 2:} Poor. Significant alteration of foreground objects (e.g., missing body parts, severe structural artifacts) or noticeably altered/unnatural motion.
\textbf{- 3:} Fair. Minor changes to the foreground objects (e.g., slight blurring of internal details, small localized artifacts) or slight jitter in their original motion.
\textbf{- 4:} Good. Foreground objects and their motion are almost perfectly preserved, with only barely noticeable minute artifacts upon close inspection.
\textbf{- 5:} Perfect. All non-background elements and their original motion remain exactly as they were in the source video. \\
\textbf{The third score: Overall Visual Quality} \\
\textbf{Objective:} Evaluates the synthesis of Visual \& Temporal Seamlessness (Edge, Blend \& Stability) and Physical Consistency (Lighting, Perspective, Motion \& Depth) of the video after editing. CRITICAL EXCEPTION: Do NOT deduct points for visual quality issues, noise, or artifacts that were already present in the original, unedited video. \\
\textbf{- 1:} Fail. Severe composite errors (large tearing, flickering, jittering edges) OR severe physical mismatch (conflicting light, floating subject, wrong horizon, no parallax during camera motion). Obvious fake at a glance.
\textbf{- 2:} Poor. Clear cut-out halos, obvious edge 'boiling', or noticeable inconsistencies in lighting, scale, and perspective shifts during motion.
\textbf{- 3:} Fair. Acceptable overall but visible flaws on closer inspection: slight edge blur, minor temporal shimmer, or small errors in shadows, depth, and perspective tracking.
\textbf{- 4:} Good. Nearly invisible seams; edges are stable, lighting/scale/depth are well-matched, and perspective tracks convincingly with camera motion. Only minor issues visible.
\textbf{- 5:} Perfect. Flawless composite and physical realism. Edges, light, shadows, perspective, and atmospheric depth are perfectly coherent and temporally stable throughout the video. \\

\textbf{Example Response Format:}
You are required to return a dictionary structured as follows: \{"Condition Preserving": [A number from 1 to 5], "Instruction Following": [A number from 1 to 5], "Overall Visual Quality": [A number from 1 to 5]\}. \\

{\bfseries\color{red}\verb=<|user_input|>=}\\

\end{AcademicBox}
\vspace{-1em}
\caption{Evaluation prompt template for TIV2V task (background change).\looseness=-1}
\label{fig:bench_tiv2v_back}
\end{figure}

\begin{figure}[t]

\begin{AcademicBox}[ Evaluation Prompt Template for TIV2V Task (local object replacement)]
You are an expert data rater specializing in grading video object replacement edits. You will be provided with an original video, a reference image, the edited video, and the corresponding editing instructions. The instructions dictate replacing a specific object in the original video with a subject from the reference image. Your task is to evaluate the editing performance on a 5-point scale across three key dimensions, paying close attention to temporal consistency. \\
\textbf{The first score: Instruction Following} \\
\textbf{Objective:} Evaluates whether the model successfully executed the editing instruction (e.g., replacing the correct target object with the specified new object) for the correct duration and at the correct position, and whether the newly added object naturally integrates into the original scene rather than looking like a rigid insertion. \\
\textbf{- 1:} Fail. The target was not replaced, or a completely unrelated object/region of the video was edited.
\textbf{- 2:} Poor. Attempted replacement, but the instruction was poorly executed (e.g., replacement only happens in a few frames, or the edit occurs in the wrong spatial location).
\textbf{- 3:} Fair. The basic instruction was followed (target replaced), but with noticeable errors in execution (e.g., incorrect count, wrong pose/scale, or failure to maintain the replacement consistently across all frames), or the new object feels stiffly, abruptly, or unnaturally forced into the scene.
\textbf{- 4:} Good. The instruction was followed correctly for the entire duration. The correct target was replaced with accurate placement and scale, with only minor deviations. And it integrates almost naturally into the original scene's context, with only minor stiffness or slight contextual mismatch.
\textbf{- 5:} Perfect. Flawless execution. All specified objects were replaced perfectly. The new object fits so naturally into the scene's context that it feels exactly as if it originally belonged there, fulfilling the prompt perfectly. \\
\textbf{The second score: Condition Preserving} \\
\textbf{Objective:} Evaluates two critical preservation aspects: (A) How well the newly added object retains the identity and intricate details of the subject in the reference image, and (B) How perfectly the unedited regions, non-target objects, and the original temporal dynamics (e.g., original background motion, natural evolving lighting, camera shifts) from the original video are preserved. \\
\textbf{- 1:} Fail. The new object bears no resemblance to the reference image, OR the unedited regions/temporal dynamics are completely destroyed or frozen.
\textbf{- 2:} Poor. Major identity loss in the new object, OR significant warping, distortion, or loss of original temporal motion in the unedited regions.
\textbf{- 3:} Fair. The new object captures the general idea but has obvious attribute discrepancies, OR there are visible, distracting changes to the unedited regions, such as background elements losing their original natural movement or temporal flow.
\textbf{- 4:} Good. The new object closely matches the reference image with only minor attribute errors, AND the unedited regions almost preserve their original appearance and temporal dynamics, with minor changes.
\textbf{- 5:} Perfect. Flawless preservation. The new object exactly matches the reference subject's details, AND absolutely zero changes occurred in the unedited areas, e.g., the original background and its temporal changes/motion are perfectly intact. \\
\textbf{The third score: Overall Visual Quality} \\
\textbf{Objective:} A comprehensive evaluation of Visual Naturalness, Temporal Stability, Physical/Motion Integrity, and how seamlessly the new object blends into the scene (lighting, shadows, perspective). CRITICAL EXCEPTION: Do NOT deduct points for visual quality issues, noise, or artifacts that were already present in the original, unedited video. \\
\textbf{- 1:} Fail. Severe visual degradation introduced by the edit. Video heavily broken or the new object is severely deformed, flickers uncontrollably, or floats/slides unnaturally (complete motion tracking failure).
\textbf{- 2:} Poor. Obvious artifacts introduced. Noticeable flickering seams, missing/static shadows, poor occlusion, or the new object's motion clearly disconnects from the camera/scene movement. Fails to blend into the environment.
\textbf{- 3:} Fair. Acceptable quality but with visible temporal or physical inconsistencies introduced by the edit. Minor flickering, edge fuzzying, lighting shifts over time, or small tolerable drifts in motion tracking. Integration is passable but imperfect.
\textbf{- 4:} Good. High visual quality and stable style. The new object's motion is well-tracked, interacts realistically with the scene (accurate shadows/reflections), and naturally blends in. Only Tiny temporal artifacts are visible.
\textbf{- 5:} Perfect. Completely seamless, temporally stable, and dynamically flawless. Perfect motion tracking, perspective, and lighting integration. The new object blends perfectly into the scene in every frame, making the edit completely undetectable to a casual viewer. \\

\textbf{Example Response Format:}
You are required to return a dictionary structured as follows: \{"Condition Preserving": [A number from 1 to 5], "Instruction Following": [A number from 1 to 5], "Overall Visual Quality": [A number from 1 to 5]\}. \\

{\bfseries\color{red}\verb=<|user_input|>=}\\

\end{AcademicBox}
\vspace{-1em}
\caption{Evaluation prompt template for TIV2V task (local object replacement).\looseness=-1}
\label{fig:bench_tiv2v_change}
\end{figure}
\end{document}